\documentclass[10pt,twocolumn,letterpaper]{article}

\usepackage[pagenumbers]{iccv} % To force page numbers, e.g. for an arXiv version

\definecolor{iccvblue}{rgb}{0.21,0.49,0.74}
\usepackage[pagebackref,breaklinks,colorlinks,allcolors=iccvblue]{hyperref}

\usepackage[accsupp]{axessibility}  % Improves PDF readability for those with disabilities.
\usepackage{booktabs}
\usepackage{multirow}
\usepackage[normalem]{ulem}
\useunder{\uline}{\ul}{}

\def\paperID{3001} % *** Enter the Paper ID here
\def\confName{ICCV}
\def\confYear{2025}

\title{SketchSplat: 3D Edge Reconstruction via
Differentiable \\ Multi-view Sketch Splatting}

\author{
        Haiyang Ying,\ \ \ 
        Matthias Zwicker\\
        University of Maryland, College Park \\
        \tt\small\href{https://oceanying.github.io/SketchSplat}{https://oceanying.github.io/SketchSplat}
}%

\usepackage{xcolor}
\newcommand{\haiyang}[1]{\textcolor{black}{#1}}

\begin{document}
\maketitle
\begin{abstract}

Edges are one of the most basic parametric primitives to describe structural information in 3D. 
In this paper, we study parametric 3D edge reconstruction from calibrated multi-view images. 
Previous methods usually reconstruct a 3D edge point set from multi-view 2D edge images, and then fit 3D edges to the point set.
However, noise in the point set may cause gaps among fitted edges, and the recovered edges may not align with input multi-view images since the edge fitting depends only on the reconstructed 3D point set.
To mitigate these problems, we propose \textbf{SketchSplat}, a method to reconstruct accurate, complete, and compact 3D edges via differentiable multi-view sketch splatting. We represent 3D edges as sketches, which are parametric lines and curves defined by attributes including control points, scales, and opacity. During reconstruction, we iteratively sample Gaussian points from a set of sketches and rasterize the Gaussians onto 2D edge images. 
% Then the gradient of the image error with respect to the input 2D edge images can be back-propagated to optimize the sketch attributes. 
Then the gradient of the image loss can be back-propagated to optimize the sketch attributes. 
Our method bridges 2D edge images and 3D edges in a differentiable manner, which ensures that 3D edges align well with 2D images and leads to accurate and complete results. 
%
% We also propose a series of adaptive topological operations and apply them along with the sketch optimization. The topological operations help reduce the number of sketches required while ensuring high accuracy, yielding a more compact reconstruction.
We also propose a series of adaptive topological operations to reduce redundant edges and apply them along with the sketch optimization, yielding a more compact reconstruction.
Finally, we contribute an accurate 2D edge detector that improves the performance of both ours and existing methods.
Experiments show that our method achieves state-of-the-art accuracy, completeness, and compactness on a benchmark CAD dataset.

\end{abstract}
\section{Introduction}
\label{sec:intro}

% What is the problem?
% Why is it interesting and important?
% What others have done?
% Why are they not satisfactory?
% What is your idea
% Why it is better?

\begin{figure}[thb]
  \centering
  % \fbox{\rule{0pt}{2in} \rule{0.9\linewidth}{0pt}}
   \includegraphics[width=1.0\linewidth]{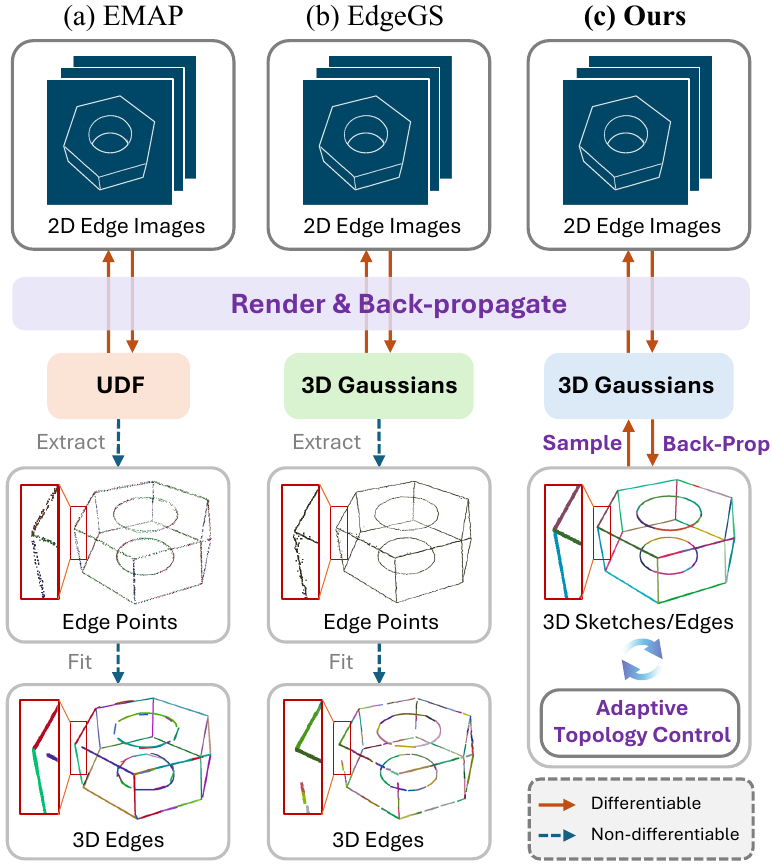}

   \caption{We propose \textbf{SketchSplat}, a multi-view 3D edge reconstruction method based on differentiable sketch splatting. 
   Previous methods~\cite{li2024emap, chelani2024edgegs} first recover edge points and then fit them to 3D edges, which is sensitive to noise and often leads to incomplete results. Instead, we sample Gaussian points from 3D sketches and rasterize them to compute a rendering loss, enabling direct edge parameter optimization under image supervision. Additionally, we design adaptive topology control methods to refine edge structures during training and contribute an improved 2D edge detection method for more accurate reconstruction.}
   \label{fig:teaser}
\end{figure}

% Different from point-based representations, continuous edges provide a more compact way to describe and understand the structure of the 3D world. 
% As one of the most fundamental elements of visual perception, edge modeling is crucial for understanding the geometry and structure of the 3D world.
% Edge-based representation also aligns closely with human perception of the 3D world. 
% 3D edge reconstruction is a fundamental problem in computer vision and graphics, facilitating downstream applications in CAD modeling\cite{choi2023app_cad_1, zhang2023app_cad_2, li2024app_cad_3}, SLAM\cite{smith2006lineslam, hruby2024handposeestimation, liu1990camlocalization, wu2020edgeodometry}, and autonomous driving\cite{qiao2023lanemapping, qin2021lightautonomousdriving, cheng2021roadmapping}. 

Edges serve as one of the most fundamental elements of visual perception, playing a crucial role in understanding the geometry and structure of the 3D world. Unlike point-based representations, continuous edges offer a more compact and perceptually aligned way to describe 3D structures. As a cornerstone of computer vision and graphics, 3D edge reconstruction underpins a wide range of applications, including CAD modeling \cite{choi2023app_cad_1, zhang2023app_cad_2, li2024app_cad_3}, SLAM \cite{smith2006lineslam, hruby2024handposeestimation, liu1990camlocalization, wu2020edgeodometry}, and autonomous driving \cite{qiao2023lanemapping, qin2021lightautonomousdriving, cheng2021roadmapping}.
Despite its significance, achieving high-quality 3D edge reconstructions remains a challenging task due to the inherent difficulty of ensuring the accuracy and completeness of reconstructed edges and maintaining the connectivity of the reconstructed structure.
% due to the discrete property of parametric edges and 

Traditional methods \cite{hofer2014sparseline3drecon, bignoli2018mvs3dedgerecon, wei2022elsr, liu2023limap} usually detect 2D line segments in each view. Then the 2D lines are lifted to 3D via line matching and triangulation across views. While these methods achieve high accuracy and efficiency, multi-view consistent detection and robust line matching are difficult to achieve, leading to incomplete results. In addition, these methods are usually not capable of handling lines and curves simultaneously.

In recent years, differentiable rendering has emerged as a promising technique to reconstruct 3D contents for novel view synthesis \cite{mildenhall2021nerf, barron2023zipnerf, muller2022instant, ying2023parf}, surface reconstruction \cite{wang2021neus, ma2020neuralpull, li2023neuralangelo, fu2022geoneus}, and animation \cite{park2021nerfies, pumarola2021dnerf, wu20244dgs}. 
In the field of edge reconstruction, several works introduced the idea of differentiable rendering to recover edge information \cite{ye2023nef, li2024emap, chelani2024edgegs}.
Instead of detecting parametric 2D edge segments explicitly, they leverage pixel-wise 2D edge intensity images \cite{su2021pidinet, poma2020dexined} as a supervision signal and %, where more edge information can be maintained.
optimize a point-based 3D edge representation via differentiable multi-view rendering, resulting in an extracted 3D edge point set. 
% After training, the rendered 3D points should match well with multi-view 2D intensity maps. 
Parametric 3D edges are then obtained by fitting lines and curves to the extracted 3D points. 
Since the edge-fitting step relies purely on the recovered 3D edge points but not the input 2D edge images, the fitted 3D edges may not align well with the 2D edge images. In addition, noise in the 3D edge points usually leads to discontinuous and fragmented edges. 

To address these limitations, we propose \textbf{SketchSplat}, a differentiable edge reconstruction framework that directly optimizes parametric 3D edges by fitting them to the input 2D edge images using differentiable rendering. 
% Our experiments show that this leads to more accurate and complete edge reconstruction with a compact set of parametric edges that more faithfully capture the original connectivity.
Our experimental results demonstrate that this leads to more accurate, complete, and compact reconstructions, more faithfully reflecting the original connectivity.
%Our motivation is: the connectivity information in 2D images is an essential clue for keeping resulting 3D edges connected and compact. To achieve this, we propose a method to bridges 2D edge images and 3D parametric edges in a differentiable manner.
%

Firstly, we propose to model 3D edges as sketches, which are parametric lines and curves with control points, scales, and opacity. These sketches can be initialized from existing methods \cite{ye2023nef, li2024emap, chelani2024edgegs}, and can be sampled as a set of 3D Gaussian points for differentiable rendering. 
To optimize the sketches, we sample Gaussian points from the sketches and rasterize them in each view to calculate the rendering loss between the rendered and the ground-truth edge image. Then the error can be back-propagated to update the sketch attributes. 
Since we optimize our sketch parameters directly using the 2D edge images, the resulting edges will align well with the 2D edges in the input images.
To further improve the compactness of the reconstructed edges, we leverage a series of topological operations to connect, merge, and filter sketches during optimization.
Finally, we observe that one common limitation of existing methods \cite{ye2023nef, li2024emap, chelani2024edgegs} lies in the inaccuracy of 2D edge images. To mitigate this, we propose an accurate and robust edge detection method, which is validated to improve both our approach and existing methods by a large margin. 
% To handle CAD objects with smooth convex surface, we also contribute a post-filtering method, which takes advantage of multi-view depth estimation to filter out noisy silhouette edges.

Experiments demonstrate that our method achieves state-of-the-art performance on the ABC-NEF dataset \cite{ye2023nef}, surpassing previous approaches by over 10\% in F-score.
% Our approach enables end-to-end optimization of sketch parameters, reducing error accumulation and improving the connectivity of reconstructed edges. 
In summary, key innovations of our method include:

\begin{itemize}
    
    \item Differentiable Parametric Optimization: We contribute a pipeline that directly optimizes parametric edge representations via differentiable multi-view rasterization, leading to SOTA accuracy and completeness.

    \item Adaptive Topological Control: We propose a series of topological operations to further enhance edge continuity and reduce redundant edges during training, yielding more compact reconstruction results.

    \item Accurate 2D Edge Detection: We propose a more accurate 2D edge detection method, which shows large improvements for both our approach and existing methods.

    % \item Post-processing - Handling Silhouette Edges: A filtering mechanism is proposed to filter the edges that arise from smooth convex surfaces, which obtains better performance for more general CAD models.
    
\end{itemize}

\section{Related Work}
\label{sec:related_work}

%-------------------------------------------------------------------------
\paragraph{Line Matching and Triangulation.} % Matthias: to save space

Conventional approaches reconstruct 3D lines \cite{bartoli2005sfmline_tra1, baillard1999automatic_tra2, chandraker2009movinginstereo_tra3, micusik2017sfmline_tra4, hofer2014sparseline3drecon, bignoli2018mvs3dedgerecon} or curves \cite{kahl2003multiview_trac1, kaminski2001multiple_trac2, robert1991curve_trac3} by enforcing multi-view geometric constraints. Given detected 2D edge segments and their descriptors, they match the line descriptors across multiple views and perform triangulation to lift 2D edges into 3D. However, these methods face challenges in consistently detecting edges and robustly matching lines, especially when occlusions cause endpoint inconsistencies across views.
Despite notable advancements by involving structural \cite{wei2022elsr, schindler2006manhattan} or learning-based priors \cite{liu2023limap, pautrat2023deeplsd, pautrat2021sold2, pautrat2023gluestick}, line detection and matching remain critical bottlenecks, limiting overall performance.
While most of the methods can only handle a single type of edges, in this paper we aim at recovering both lines and curves.

%-------------------------------------------------------------------------
\paragraph{Differentiable Rendering Methods.}

The advancement of differentiable rendering opens up a new way for 3D reconstruction \cite{mildenhall2021nerf, wang2021neus, li2023neuralangelo}.
% makes it possible to bridge 3D edges and 2D edge images.
Inspired by NeRF~\cite{mildenhall2021nerf}, NEF~\cite{ye2023nef} proposes to optimize an implicit neural radiance field with a multi-view rendering loss. After training, 3D edge points are extracted from the radiance field and curves are fitted to these points.
NEAT~\cite{xue2024neat} builds on VolSDF~\cite{yariv2021volsdf} and uses a signed distance field to represent both object surfaces and endpoints of 3D lines. The surface-based modeling approach restricts this method to objects with rich texture.
EMAP~\cite{li2024emap} adopts an unsigned distance function (UDF), where sampled points can be displaced towards the zero-level set of the UDF to be more accurate.
While methods based on neural field are time consuming to optimize, EdgeGS~\cite{chelani2024edgegs} represents 3D edge fields with explicit Gaussian points and leverages Gaussian splatting~\cite{kerbl20233dgs} to render and optimize Gaussian points, leading to more efficient edge reconstruction.
A common issue of these methods is that the edge fitting process relies only on the extracted 3D edge points, hence noise in the reconstructed 3D  points may lead to inaccurate and incomplete edges.
% Instead we represent 3D edges as sketches that are defined parametrically using control points, scales and opacity, and directly optimize the sketch parameters using differentiable rendering. Differentiable rendering uses sketch sampling and rasterization to align the 3D edges with 2D edge images directly, yielding more accurate and complete results.
Instead, we represent 3D edges as sketches defined parametrically by control points, scales, and opacity, and directly optimize these parameters via differentiable rendering. By leveraging sketch sampling and rasterization, our differentiable rendering aligns 3D edges with 2D edge images, resulting in more accurate and complete reconstructions.

%-------------------------------------------------------------------------
\paragraph{Differentiable Vector Graphics Optimization.}

DiffVG~\cite{Li:2020:DiffVG} proposes a differentiable rasterizer that can optimize 2D vector graphics with 2D image supervision in a differentiable manner. However, the differentiation process relies on distance measurements between the rasterized curves and pixels, which needs to be custom tailored for each type of curve. 
3Doodle~\cite{choi20243doodle} extends the idea of DiffVG~\cite{Li:2020:DiffVG} to reconstruct 3D curve-based object abstractions using multiple input views. They optimize a set of B\'ezier curves by minimizing the perceptual loss between the rasterized image and multi-view CLIP feature maps \cite{radford2021clip, vinker2022clipasso}. Since CLIP features only contain abstract information, 3Doodle is not suitable for recovering accurate edges of objects like CAD models. Moreover, they approximate the projection of B\'ezier curves using an orthographic camera model.
%
% Neural 3D Strokes\cite{duan2024strokenerf} propose to represent the 3D object with a combination of strokes. Each stroke is a signed distance field defined by the stroke parameters, which can be optimized via differentiable rendering. However, they focus on rendering quality instead of meaningful and accurate geometric structure recovery.
%
Instead, we get inspiration from 3DGS~\cite{kerbl20233dgs} and propose to optimize 3D sketches via sampling and rasterization of Gaussian points sampled from sketch parameters, which can be easily extended to different types of curves and achieves accurate 3D edge reconstruction.

\section{Preliminaries: 3D Gaussian Splatting}

%-------------------------------------------------------------------------
%\paragraph{3D Gaussian Splatting}

We leverage 3D Gaussian splatting (3DGS)~\cite{kerbl20233dgs} for initialization and differentiable sketch rendering. 3DGS proposes to model 3D scenes with a collection of anisotropic 3D Gaussian points. Each  point is defined as a Gaussian distribution $G(x)=e^{- \frac{1}{2} (x-\mu)^T \Sigma^{-1} (x - \mu)}$, where $\mu\in R^3$ is the center position or mean, and $\Sigma\in R^{3\times3}$ is the covariance matrix. Each Gaussian has an opacity $o \in R$ and a view-dependent color value $c(d)=SH(d)\in R^3$ modeled with spherical harmonics.
Following Zwicker et al.~\cite{zwicker2001ewasplatting}, the Gaussians can be rendered to 2D images using an efficient volume rendering approach, where the color for each camera ray $r$ is 
\begin{equation}
    c(r)=\sum_{i\in N}c_i(d(r)) \alpha_i \prod_{j=1}^{i-1} (1-\alpha_j), \alpha_i=o_i g_i(r),
    \label{eq:rasterization}
\end{equation}
where $g_i(r)$ is the line integral of the 3D Gaussian $G_i$ along ray $r$, and the indices $i$ are ordered in front to back manner. By calculating the gradient of a loss $\mathcal{L}$ between the rendered and  ground truth image, the parameters of all Gaussians can be optimized using gradient descent. A typical loss function is a combination of an ${L}_1$ and a perceptual loss such as
\begin{equation}
    \mathcal{L}=\lambda \mathcal{L}_1 + (1-\lambda) \mathcal{L}_{D-SSIM}.
\end{equation}
Given input images, the initial Gaussians are usually obtained from SfM~\cite{schonberger2016structure} methods. During training, several operations are adopted to adaptively control the density and distribution of Gaussians. For example, Gaussians with low opacity will be culled, and Gaussians with high gradient will be duplicated or split to densify specific regions.

In SketchSplat, we leverage 3DGS~\cite{kerbl20233dgs} as an intermediary to bridge 2D edge images and 3D parametric edges, enabling the optimization of 3D edges through differentiable image rendering.

\section{Method}
\label{sec:method}

\begin{figure*}[thb]
  \centering
  % \fbox{\rule{0pt}{2in} \rule{0.9\linewidth}{0pt}}
   \includegraphics[width=1.0\linewidth]{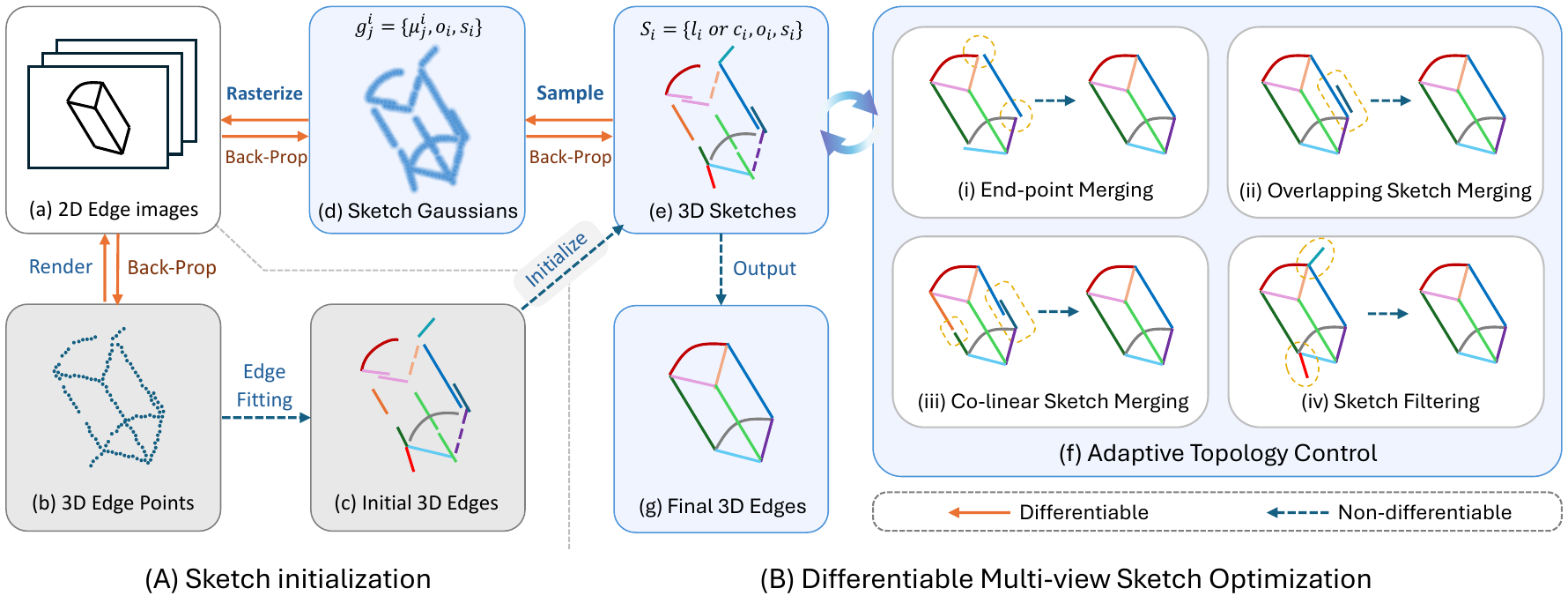}

   \caption{
   % \matthias{Implement suggestions in pipeline figure. The figure should appear on the previous page where Section 4 starts.} 
   \textbf{Overview of SketchSplat pipeline}: (a) We extract edge images using a novel approach (Sec.~\ref{sec:edge-image-extraction}). (b) We then obtain initial edge points using EdgeGS~\cite{chelani2024edgegs} and (c) fit parametric sketches defined as lines and third-order B\'ezier curves. The core step of our approach is to optimize the sketches by (d) sampling and rasterizing them in a differentiable manner and back-propagating gradients of the image loss to (e) sketch parameters. In addition, during training we apply (f) a set of topological operations to improve compactness and connectivity of the reconstructed sketches. (g) After training, we take the optimized geometric parameters of sketches as the reconstructed 3D edges.}
   \label{fig:pipeline}
\end{figure*}

Given multi-view posed RGB images as input, our target is to recover the parametric edges for the 3D objects. Fig.~\ref{fig:pipeline} provides an overview of our approach.
We firstly compute an edge image for each RGB image using our proposed 2D edge detector.
Then we represent edges as 3D parametric sketches, including both lines $l\in \mathbb{R}^{2\times3}$ and third-order B\'ezier curves with four 3D control points $c\in \mathbb{R}^{4\times3}$. 
We initialize a set of sketches from existing edge reconstruction methods,
% (including LIMAP, NEF, EMAP, EdgeGS). 
and treat their control points, along with per-sketch scales and opacity, as optimizable parameters.
During optimization, we use a sample-based method to convert sketches to Gaussian points, which can be rasterized onto 2D images to calculate the image loss. We back-propagate the gradient of the loss to optimize the sketch parameters. 
% using gradient descent.
%
In addition, we introduce a set of topological operations to adaptively refine the sketches during training, yielding more accurate and compact edge reconstruction results.

%-------------------------------------------------------------------------
\subsection{Edge Image Extraction}
\label{sec:edge-image-extraction}

Previously, some methods~\cite{liu2023limap, xue2024neat} extract edges with line detectors like LSD~\cite{von2008lsd} and SOLD2~\cite{pautrat2021sold2}. While these methods produce accurate parametric lines in 2D, they may fail to handle curve regions and result in incomplete results. Recent methods \cite{ye2023nef, li2024emap, chelani2024edgegs} use neural network-based edge detectors \cite{su2021pidinet, poma2020dexined} to detect edge intensity maps as supervision signal. However, since these detectors are not trained specifically on CAD-like objects, they may miss some significant edges in 2D or wrongly recover shading textures as real edges. Finally, we observed that these detectors introduce bias by consistently shifting detected edges slightly from certain real edges, which leads to multi-view inconsistency in 3D edge reconstruction (see supplementary). We address these issues by proposing a new edge detection method, which leverages 
geometric cues (e.g., per-pixel depth and normal estimation of 2D RGB images) to produce more accurate edge detection results.
% the fact that our problem statement assumes the availability of multiple input images.

For each RGB image, we combine information from three parts: foreground mask $A$, depth map $D$, and normal map $N$. The foreground mask can be extracted using the image alpha-channel (if available), a thresholding operation or possibly using semantic segmentation, providing object boundary information. We obtain a depth image for each view by reconstructing the scene with 2DGS~\cite{huang20242dgs} and rendering depth maps from each view. We use a normal estimator~\cite{ye2024stablenormal} to generate normal maps $N$ for each view. 
Then we calculate gradient maps $g(D)$ and $g(N)$ with the Sobel operator~\cite{sobel1968sobel}, threshold them with $t_d$ and $t_n$, and combine them via the OR operation, followed by a Gaussian filter $G_f(\cdot)$ to smooth the boundaries of the edges. In summary, we call our edge detector 2DGS-SN, defined as 
\begin{equation}
    E = G_f * (A|(g(D) > t_d)| (g(N)>t_n)).
    \label{equ:edge_detector}
\end{equation}
In Sec.~\ref{sec:experiments} we show that this simple approach improves existing methods~\cite{chelani2024edgegs} on CAD-like objects by a large margin.

% The advantage of this method is that we fully utilize the RGB images to generate high-quality edge images.

%-------------------------------------------------------------------------
\subsection{Sketch Splatting}

\paragraph{Representation.}
We represent edges as 3D parametric sketches, including both lines $l\in \mathbb{R}^{2\times3}$ and third-order B\'ezier curves with control points $c\in \mathbb{R}^{4\times3}$.
This is the same as the output format of EdgeGS\cite{chelani2024edgegs}. While EdgeGS optimizes 3D point positions first and then fits 3D lines and curves to the 3D points, we optimize the parameters of lines and curves directly via multi-view differentiable rasterization. In addition to the optimizable geometry parameters ($l$ or $c$), we equip each sketch with an optimizable opacity $o \in \mathbb{R}$ and a local scale $s \in \mathbb{R}^3$ to model the sketch intensity and thickness, respectively. These attributes will support Gaussian sampling and rasterization in the next steps.

\paragraph{Sketch Initialization.}
Unlike point-based methods~\cite{li2024emap, chelani2024edgegs}, it is non-trivial to initialize sketches directly from multi-view edge images. Instead, inspired by 3DGS~\cite{kerbl20233dgs}, we first initialize sketches with existing point-based methods like EdgeGS~\cite{chelani2024edgegs}. By default, we select EdgeGS because it provides a sufficient number of short 3D edges to cover the objects, and achieves the highest completeness among previous methods, despite some fragmentation and noise.

% \paragraph{Sketch Optimization using Differentiable Rendering via Gaussian Sampling and Splatting.}
\paragraph{Differentiable Sketch Optimization.}

Our key idea is to bridge the gap between 2D edge images and 3D sketches via differentiable rendering and gradient descent optimization, which significantly improves sketch reconstruction over initialization method EdgeGS~\cite{chelani2024edgegs}.  Inspired by 3DGS~\cite{kerbl20233dgs} and DiffVG~\cite{Li:2020:DiffVG}, we choose 3D Gaussians as an intermediate representation to achieve differentiable sketch rendering.
% Given a set of sketches as initialization, we will sample points on sketches as gaussian points in 3DGS\cite{kerbl20233dgs}. 

For each gradient descent step, we first sample points from each sketch with a constant step size (5mm for all experiments), resulting in a point set $P$. Each point will be assigned the opacity and scale value of the corresponding sketch. To model the orientation, the main direction of each point is sampled as the local tangent direction of the sketch. 
Then we follow 3DGS\cite{kerbl20233dgs} to rasterize the Gaussian points onto a chosen 2D view to generate a rendered edge image $E^*$. We calculate the $L_1$ loss between $E^*$ and the ground truth edge image $E_{gt}$, and back-propagate the gradient to update the sketch parameters $S_i = \{l_i \ \text{or}\  c_i, o_i, s_i\}$.  We follow EdgeGS~\cite{chelani2024edgegs} to sample an equal number of foreground and background pixels for $L_1$ loss calculation.
% \matthias{Add a sentence about the termination criterion. Is it a fixed number of steps?} \haiyang{Yes, it has a fixed training steps.}
%
%After projecting the sampled Gaussian points onto a 2D edge image, if the rendered sketch is not sufficient to cover the neighboring real edge, the sketch should be transformed to cover larger edge regions, yielding lower rendering loss. 
%

%-------------------------------------------------------------------------
\subsection{Adaptive Topology Control}

A good edge reconstruction, which could be used for CAD modeling for example, should not only accurately match the 2D edge images, but also provide a compact, well-connected set of edges. Neighboring edges should be connected correctly, overlapping sketches should be merged, and the number of sketches should be as small as possible while ensuring accurate matching of 2D edge images. For this purpose we deploy topology control operations adaptively during optimization.
While some existing methods~\cite{chelani2024edgegs} do not consider topology optimization, others~\cite{ye2023nef, li2024emap} try to connect and merge neighboring sketches in the post-processing stage. However, this may be sensitive to noise in the extracted 3D point set and cannot ensure that the resulting edges are still consistent with the input multi-view 2D edge images.
Instead, since our method directly optimizes sketch parameters using iterative gradient descent, we can execute topological operations along with the multi-view optimization steps, as shown in Fig.~\ref{fig:pipeline}(f).

\paragraph{End-point Merging.} 
In our representation, each sketch has two end-points. If there exist two end-points, one from sketch $S_i$ and one from sketch $S_j$, that are sufficiently close (i.e. the point distance is smaller than threshold $th_{connect}=10mm$), we will connect sketch $S_i$ and $S_j$ by merging those two end-points into one point (see Fig.~\ref{fig:pipeline}(f-i)). However, this means the number of optimizable points could differ from the number of control points $|\{l\}|*2+|\{c\}|*4$. To accommodate this, we maintain an optimizable point set $P$, and the control points of each line $l$ and curve $c$ are represented by indexing point set $P$. This representation also supports our additional topological operations.

\paragraph{Overlapping Sketch Merging.}
As shown in Fig.~\ref{fig:pipeline}(f-ii), if one sketch is almost entirely covered by another, the smaller one will be merged into the larger one. Specifically, for sketch $S_i$ and $S_j$, we sample two point sets $P_i$ and $P_j$ and calculate the distance between each pair of points. Then for $P_i$, we calculate the ratio $r_{i\rightarrow j}$ of points that have close neighboring points in $P_j$, where the neighbor threshold is $th_{neighbor}=10mm$. If $r_{i\rightarrow j} > th_{overlap}=80\%$, sketch $S_i$ will be merged in sketch $S_j$. If both $r_{i\rightarrow j}$ and $r_{j\rightarrow i}$ surpass $th_{overlap}$, we will choose the sketch with larger overlapping ratio to be merged.

\paragraph{Co-linear Sketch Merging.}
If two lines are co-linear and nearly connected, they should be merged into a longer line (see Fig.~\ref{fig:pipeline}(f-iii)). 
The co-linear condition is: two lines should have similar direction ($th_{dir} = 5^\circ$), and the maximum of the projective distance/offset should be small enough ($th_{offset} = 10mm$). In addition, to prevent gaps between two lines, one line will be projected onto the other one to calculate if there is a gap and if the gap is small enough ($th_{connect}$). To merge the lines, we take the two end-points with the furthest distance after projection and delete the other two endpoints from point set $P$. Additionally, to accelerate this procedure, we use the axis-aligned bounding box of each sketch to quickly filter out all the sketch pairs that cannot meet the merging criterion.

% \paragraph{Add New sketches.}
% The initialized sketches may not cover the whole object. To achieve a higher completeness, we adaptively add new sketches into the scene. New sketches can be initialize from Gaussian points $P_{EdgeGS}$ reconstructed from EdgeGS\cite{chelani2024edgegs} (before edge fitting). Specifically, we evaluate the K nearest neighbor ($k=5$) of point set $P_{EdgeGS} \cup G$, where $G$ is the point set sampled from our sketches. For each point in $P_{EdgeGS}$, if it does not have a neighbor in $G$, it will be take as a under-reconstructed point. For all these points, we will choose part of them and initialize new sketches with their position and direction vector.

\paragraph{Sketch Filtering.}
% Follow 3DGS\cite{kerbl20233dgs}, we temporally filter out sketches with low opacity ($th_{opacity}=0.08$). 
After training, we adopt a multi-view filter method to filter sketches that are not visible. We borrow the refinement idea from EMAP~\cite{li2024emap}. 3D points can be projected onto 2D edge images to see if they are aligned with the 2D edges by checking the edge intensity of the projected pixel position. A point will be labeled as invisible if it is invisible in more than $90\%$ of the views. For each sketch, we will evaluate the visibility for all of the points sampled from the sketch. If more than $th_{vis}=50\%$ of the sampled points are invisible points, the sketch will be removed. When rasterizing sketches, some sketches that should not appear due to occlusion still appear in the rendered image, which sometimes leads to incorrect reconstruction. This filter scheme will help filter out such outlier sketches.
%
% Besides, for CAD dataset like ABC-NEF\cite{koch2019abc}, we will take advantage of multi-view foreground (silhouette) masks to filter outlier sketches. The sketch will be filtered if there exists one view, in which more than $60\%$ sampled points (sampled from the sketch) locate outside the foreground mask.

% \paragraph{Post Filtering.}
% Optionally, we propose a novel edge filter method, which can help filter edges that comes from inconsistent observations of smooth convex surface (e.g. cylinders and balls). We can seek help from multi-view depth information recovered by 2DGS\cite{huang20242dgs}. 
% Specifically, for each view, we firstly use rendered 2DGS depth map to filter the invisible points sampled from our sketches (all the points behind the depth map to a margin will be excluded).
% Our principle is that a valid sketch should be observed in at least one image, in which the projected 2D sketch locates at non-steep region (i.e. the observed depth variance is not too high). We do this test for each sampled points. For each sketch, if more than $th_{post}=80\%$ of its sampled points cannot pass this test, the sketch will be removed.

\subsection{Implementation Details}
Unless otherwise specified, we initialize with EdgeGS~\cite{chelani2024edgegs}, which supplies a sufficient number of 3D edges to capture object structure.
For each scene, we optimize the sketches for 1000 epochs \haiyang{with Adam}. We update parameters only once for each epoch, where the loss is accumulated across all the training views.
All of the scenes are scaled to fit within a $1\text{m}^3$ bounding box, and topology operation hyperparameters are set accordingly.
Distance thresholds (e.g., $th_{connect}$) are set based on the desired precision and we set all of them as $1\%$ of the scene size ($1 m^3$).
The percentage and angular thresholds (e.g., $th_{overlap}$ and $th_{dir}$) work across different scenes without adjustment.
All our experiments are implemented on a single RTX A5000 NVIDIA GPU. In average, each scene in ABC-NEF~\cite{ye2023nef} dataset costs about 10 minutes for training, where both initialization and sketch optimization takes about 5 minutes, respectively.

\section{Experiments}
\label{sec:experiments}

% Please add the following required packages to your document preamble:
% \usepackage{booktabs}
% \usepackage{multirow}
% \usepackage[normalem]{ulem}
% \useunder{\uline}{\ul}{}
\begin{table*}[thb]
\centering
\small
\begin{tabular}{l|l|l|ll|lll|lll|lll}
\hline
Method                        & Detector & Modal & A$\downarrow$ & C$\downarrow$ & R5$\uparrow$  & R10$\uparrow$ & R20$\uparrow$ & P5$\uparrow$  & P10$\uparrow$ & P20$\uparrow$ & F5$\uparrow$  & F10$\uparrow$ & F20$\uparrow$ \\ \hline
\multirow{2}{*}{LIMAP\cite{liu2023limap}}        & LSD      & Line  & 9.9             & 18.7             & 36.2          & 82.3          & 87.9          & 43.0          & 87.6          & 93.9          & 39.0          & 84.3          & 90.4          \\
                              & SOLD2    & Line  & \textbf{5.9}    & 29.6             & 64.2          & 76.6          & 79.6          & {\ul 88.1}          & \textbf{96.4} & \textbf{97.9} & 72.9          & 84.0          & 86.7          \\ \hline
\multirow{3}{*}{NEF\cite{ye2023nef}}          & PiDiNeT$\dagger$  & Curve & 11.9            & 16.9             & 11.4          & 62.0          & 91.3          & 15.7          & 68.5          & 96.3          & 13.0          & 64.0          & 93.3          \\
                              & PiDiNeT  & Curve & 15.1            & 16.5             & 11.7          & 53.3          & 93.9          & 12.3          & 61.3          & 95.8          & 12.3          & 51.8          & 88.7          \\
                              & DexiNed  & Curve & 21.9            & 15.7             & 11.3          & 48.3          & 93.7          & 11.5          & 58.9          & 91.7          & 10.8          & 42.1          & 76.8          \\ \hline
\multirow{3}{*}{EMAP\cite{li2024emap}}         & PiDiNeT  & Edge  & 9.2             & 15.6             & 30.2          & 75.7          & 89.5          & 35.6          & 79.1          & 95.4          & 32.4          & 77.0          & 92.2          \\
                              & DexiNed  & Edge  & 8.8             & 8.9              & 56.4          & 88.9          & 94.8          & 62.9          & 89.9          & 95.7          & 59.1          & 88.9          & 94.9          \\
                              & 2DGS-SN  & Edge  & 8.8             & 7.9              & 63.5          & 90.9          & {\ul 96.2}          & 70.4          & 91.0          & 94.9          & 66.3          & 90.4          & 95.1          \\ \hline
\multirow{3}{*}{EdgeGS\cite{chelani2024edgegs}}       & PiDiNeT  & Edge  & 11.7            & 10.3             & 17.1          & 73.9          & 83.1          & 26.0          & 87.2          & 92.5          & 20.6          & 79.3          & 86.7          \\
                              & DexiNed  & Edge  & 9.6             & 8.4              & 42.4          & 91.7          & 95.8          & 49.1          & 94.8          & 96.3          & 45.2          & 93.7          & 95.7          \\
                              & 2DGS-SN  & Edge  & 7.4             & {\ul 7.2}        & {\ul 75.7}          & 91.0          & 96.0          & 86.9          & 95.9          & 96.4          & {\ul 80.3}          & 92.9          & 95.8          \\ \hline
\multirow{3}{*}{Ours} & PiDiNeT  & Edge  & 10.4            & 11.0             & 25.1          & 88.6          & 94.5          & 30.9          & 89.6          & 95.4          & 27.5          & 88.5          & 94.5          \\
                              & DexiNed  & Edge  & 8.5             & 8.6              & 47.5          & {\ul 93.5}    & 96.1    & 54.1          & {\ul 96.3}    & {\ul 97.1}    & 50.3          & {\ul 94.4}    & {\ul 96.3}    \\
                              & 2DGS-SN  & Edge  & {\ul 6.8}             & \textbf{5.8}     & \textbf{90.8} & \textbf{95.7} & \textbf{97.2} & \textbf{92.9} & {\ul 96.3}          & 96.6          & \textbf{91.3} & \textbf{95.4} & \textbf{96.5} \\
\hline
\end{tabular}
\caption{Quantitative evaluation on ABC-NEF dataset~\cite{ye2023nef}. Our method achieves the state-of-the-art performance on most metrics. \haiyang{A: Accuracy (mm), C: Completeness (mm), R$x$: Recall (threshold $x$ mm), P$x$: Precision (mm), F$x$: F-score (mm).}}
\label{tab:abc}
\end{table*}

\begin{figure*}[thb]
  \centering
  % \fbox{\rule{0pt}{2in} \rule{0.9\linewidth}{0pt}}
   \includegraphics[width=1.0\linewidth]{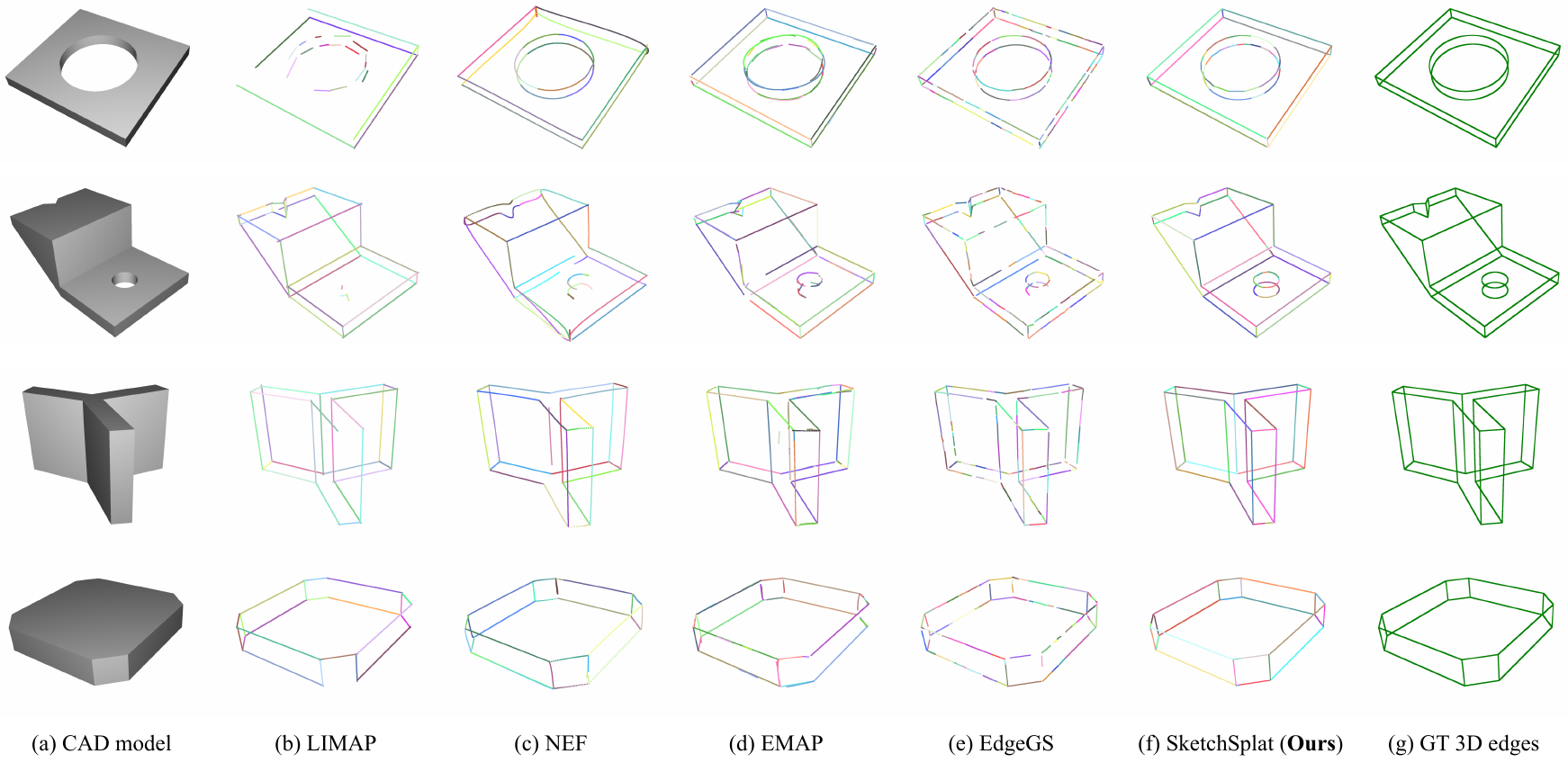}

   \caption{Qualitative comparison on the ABC-NEF dataset~\cite{ye2023nef}. Compared to the baselines, our method achieves superior accuracy and completeness, along with a favorable balance between completeness and compactness.}
   \label{fig:abc_qualitative}
\end{figure*}

%-------------------------------------------------------------------------
\subsection{Experimental Setup}

We compare our method with other multi-view edge reconstruction methods following the evaluation setup of EdgeGS~\cite{chelani2024edgegs}. The results show that our method achieves the state-of-the-art performance in both accuracy and completeness on the ABC-NEF dataset~\cite{ye2023nef,koch2019abc}. Our method also produces highly compact results with fast inference speed.

\paragraph{Datasets.}
We evaluate our method on three datasets: ABC-NEF~\cite{ye2023nef,koch2019abc} DTU~\cite{aanaes2016dtu}, and Replica~\cite{straub2019replica}. 
The ABC-NEF dataset~\cite{ye2023nef,koch2019abc} comprises 115 CAD models, each containing 50 RGB images of the target object and a parametric edge file providing ground truth 3D edges. Following EMAP~\cite{li2024emap} and EdgeGS~\cite{chelani2024edgegs}, we use 82 CAD models from ABC-NEF, excluding the ones with cylinders and spheres since smooth convex surfaces will produce silhouette edges that are not consistent in multi-views. 
For 2D edge image generation, we use PiDiNet~\cite{su2021pidinet}, DexiNed~\cite{poma2020dexined}, and our proposed 2DGS-SN as 2D edge detectors.
The DTU dataset~\cite{aanaes2016dtu} contains multi-view images of real-world scenes. We follow EdgeGS~\cite{chelani2024edgegs} and evaluate on 6 DTU scenes. We use edge images and the pseudo ground truth edges provided by EMAP~\cite{li2024emap}.
% To generate pseudo ground truth edges, EMAP  leverages the ground truth 3D surface points and filters out the points that do not project onto 2D edges. 
%
The Replica dataset~\cite{straub2019replica} consists of multiple indoor scenes. 
We follow EdgeGS~\cite{chelani2024edgegs} to conduct qualitative evaluation on Replica~\cite{straub2019replica}.

\begin{figure*}[thb]
  \centering
  % \fbox{\rule{0pt}{2in} \rule{0.9\linewidth}{0pt}}
   \includegraphics[width=1.0\linewidth]{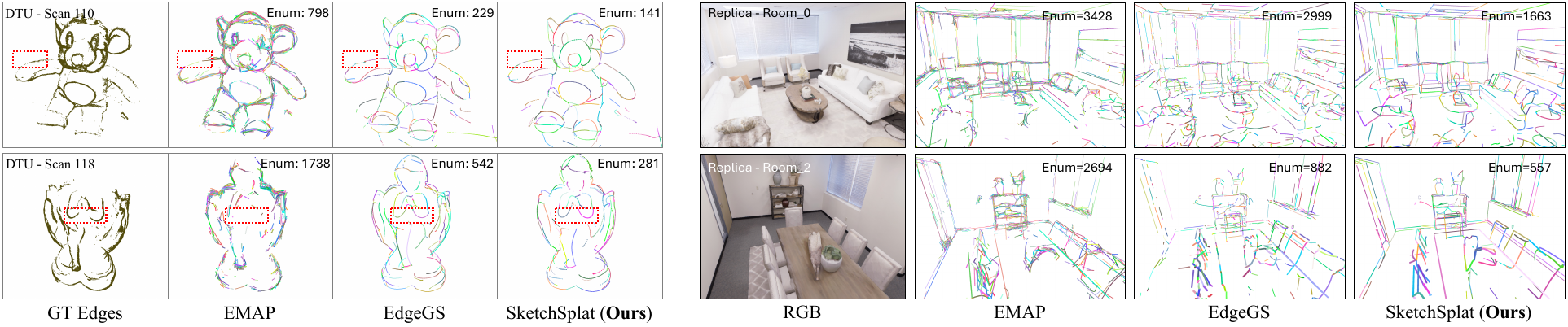}

   \caption{Qualitative comparison on the DTU and Replica datasets~\cite{aanaes2016dtu, straub2019replica}. \texttt{Enum} is the edge number and edges are visualized with varied colors. Compared to the baselines, our method achieves higher compactness and retains faithful completeness.}
   \label{fig:dtu_replica_qualitative}
\end{figure*}

\paragraph{Baselines.}

We compare our method with five baseline methods, including the state-of-the-art 3D line mapping method (LIMAP~\cite{liu2023limap}), and several  methods based on volumetric differentiable rendering (NEF~\cite{ye2023nef}, NEAT~\cite{xue2024neat}, EMAP~\cite{li2024emap}, EdgeGS~\cite{chelani2024edgegs}).
Following EMAP and EdgeGS, we do not evaluate NEAT on the ABC-NEF dataset since NEAT often fails to train for texture-less objects.

\paragraph{Metrics.}
Following the metrics used in baseline methods~\cite{li2024emap, chelani2024edgegs}, we evaluate Accuracy (A) and Completeness (C) in millimeters, and Recall (R), Precision (P), and F-score (F) as percentages at three thresholds ($5mm$, $10mm$, $20mm$). 
% All of the scenes are scaled to fit within a $1\text{m}^3$ bounding box. 
See supplementary material for more details.

%-------------------------------------------------------------------------
\subsection{Results}

\begin{figure*}[thb]
  \centering
  % \fbox{\rule{0pt}{2in} \rule{0.9\linewidth}{0pt}}
   \includegraphics[width=1.0\linewidth]{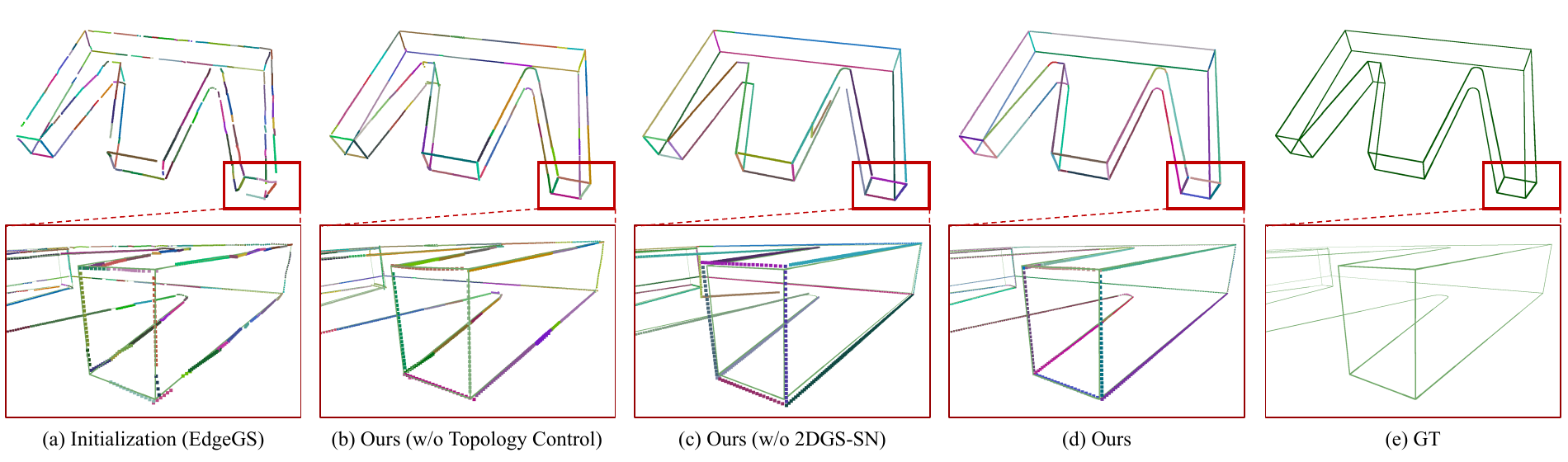}

   \caption{Ablation study on the key components of SketchSplat. Removing key components leads to noisier or redundant reconstructions.}
   \label{fig:ablation}
\end{figure*}

\paragraph{Evaluation on ABC-NEF~\cite{ye2023nef}.}
The qualitative and quantitative evaluation results on ABC-NEF dataset are shown in Fig.~\ref{fig:abc_qualitative} and Tab.~\ref{tab:abc}, respectively. 
Overall, the proposed method achieves a new state-of-the-art across most of the metrics. To the best of our knowledge, SketchSplat is the first method to achieve both accuracy (A) and completeness (C) below $7.0mm$, while also exceeding $90\%$ in recall (R5), precision (P5), and F-score (F5).

Tab.~\ref{tab:abc} shows LIMAP~\cite{liu2023limap} achieves the highest accuracy due to its robust line mapping method. However, its completeness and recall are low, as LIMAP struggles with curve reconstruction and consistent edge detection results (Fig.~\ref{fig:abc_qualitative}(b)). 
% Our method (with 2DGS-SN) achieves comparable accuracy but much higher completeness.
%
NEF~\cite{ye2023nef} produces more complete results in Fig.~\ref{fig:abc_qualitative}(c), but performs poorly on metrics as its fitting process is sensitive to noise in the extracted 3D points.
% relying solely on extracted 3D edge points.
EMAP~\cite{li2024emap} outperforms NEF with its unsigned distance field modeling and point shifting mechanism. However, Fig.~\ref{fig:abc_qualitative}(d) shows that EMAP still generates duplicated edges and misses some edges.
With PidiNeT and DexiNed as 2D detector, EdgeGS~\cite{chelani2024edgegs} surpasses EMAP in recall and precision at $10mm$ and $20mm$ thresholds, but the trend reverses for measurements under $5mm$. This is due to the thickness of 2D edges introduces bias in smaller distance evaluations. When using our detector 2DGS-SN, EdgeGS performs better than EMAP to a large margin under $5mm$ metrics.
As shown in Fig.~\ref{fig:abc_qualitative}, our SketchSplat achieves the highest precision and completeness compared to all the baselines.

As shown in Tab.\ref{tab:abc}, the proposed 2DGS-SN significantly boosts the performance of EMAP, EdgeGS, and SketchSplat, addressing the bias from PiDiNet\cite{su2021pidinet} and DexiNed~\cite{poma2020dexined}.
Notably, across all edge detectors, our method consistently achieves higher Recall, Accuracy, and F-score than EdgeGS~\cite{chelani2024edgegs}, demonstrating effectiveness of SketchSplat in aligning 3D edges with 2D edge images.

\begin{table}[t]
\centering
\begin{tabular}{c| r c}
\hline
Method                          & E-num & Time (h) \\ \hline
NEF~\cite{ye2023nef}             & 22.9       & 1:26 \\ \hline
EMAP~\cite{li2024emap}           & 45.8       & 2:30 \\ \hline
EdgeGS~\cite{chelani2024edgegs}  & 140.8      & 0:05 \\ \hline
ours                            & 44.3       & 0:10 \\ \hline
\end{tabular}
\caption{Comparison of the number of reconstructed edges \haiyang{(E-num)} and training time on ABC-NEF dataset.}
\label{tab:num_time}
\end{table}

\paragraph{Evaluation on DTU~\cite{aanaes2016dtu} and Replica~\cite{straub2019replica}.}
Fig.~\ref{fig:dtu_replica_qualitative} presents qualitative comparisons on the DTU and Replica datasets. Compared to the baselines, our method consistently produces accurate and complete results using significantly fewer edges, demonstrating a strong balance between completeness and compactness. Although the ground-truth edges in the DTU scenes are not sufficiently accurate for reliable quantitative evaluation~\cite{chelani2024edgegs}, our method still achieves metrics comparable to state-of-the-art methods EMAP and EdgeGS (see supplementary material for more details).

% Tab.~\ref{tab:dtu} shows the evaluation results on six scenes in the DTU dataset~\cite{aanaes2016dtu}. We reproduce EdgeGS~\cite{chelani2024edgegs} using its released code and configuration (denoted as EdgeGS*) to initialize sketches for our method. 
% %
% As shown in Tab.~\ref{tab:dtu}, SketchSplat without topological operations achieves comparable results to EMAP~\cite{li2024emap} and slightly outperforms EdgeGS, demonstrating its effectiveness in correcting misalignment between 3D edges and 2D edge images.
% %
% However, incorporating topological operations slightly lowers recall. This is because the merging operations aim to merge repeated edges and make the results compact, while the pseudo ground truth edges favor duplicated edges in precision (P5) and recall (R5). 

\paragraph{Speed Analysis.}
The initialization with EdgeGS costs about 5 minutes. For optimization part,
thanks to the efficiency of rasterization-based optimization, our method achieves comparable speed (about 5 mininute per scene) to EdgeGS while being significantly faster than other neural network-based approaches~\cite{ye2023nef, li2024emap}, as shown in Tab.~\ref{tab:num_time}.

\paragraph{Compactness.}
EdgeGS achieves higher recall than NEF and EMAP due to its strict edge fitting process, which produces numerous short edges, covering more edge regions as the number of edges increases. 
Instead, our method incorporates adaptive topology control to reduce redundant edges, as shown in Tab.~\ref{tab:num_time}, Fig.~\ref{fig:dtu_replica_qualitative}, and Fig.~\ref{fig:ablation}(a,b,d). This mechanism helps SketchSplat achieve higher compactness, while maintaining the highest accuracy and completeness. % in the metrics.
% In Tab.~\ref{tab:num_time}, we choose the results of each method with the best performance for comparison.

\subsection{Ablation}

\begin{table}[t]
% Ablation-v3
\centering
\small
\begin{tabular}{l|cccccr}
\hline
Method      & A$\downarrow$ & C$\downarrow$ & R5$\uparrow$ & P5$\uparrow$ & F5$\uparrow$ & E-num \\ \hline
Ours        & 6.8             & 5.8              & 90.8         & 92.9         & 91.3         & 44.28      \\ \hline
w/o 2DGS-SN & 8.5             & 8.6              & 47.5         & 54.1         & 50.3         & 33.52      \\
w/o Merge   & 6.6             & 5.7              & 91.8         & 92.7         & 91.7         & 140.72      \\
w/o Filter  & 6.8             & 5.8              & 90.9         & 92.9         & 91.3         & 44.37     \\ 
% \hline
\hline
\end{tabular}
\caption{Ablation on the proposed components using ABC-NEF. \texttt{w/o 2DGS-SN} denotes SketchSplat trained with edge images detected by DexiNed, while the others are trained with 2DGS-SN.}
\label{tab:ablation-contribution}
\end{table}

\paragraph{Key Components.}
We use the ABC-NEF dataset to conduct an ablation study on the key contributions of our method, with results presented in Fig.~\ref{fig:ablation} and Tab.~\ref{tab:ablation-contribution}.
Fig.~\ref{fig:ablation}(b,d) shows that without topology control, noisy edge segments appear around the ground truth edges, with some extending beyond the object boundary. Incorporating topology control, particularly merging operations, effectively reduces the required edge count while maintaining comparable performance on other metrics (see Tab.~\ref{tab:ablation-contribution}).
In addition, compared to edges reconstructed using DexiNed 2D edge images (Fig.~\ref{fig:ablation}(c)), our proposed 2DGS-SN better aligns edges with the ground truth (Fig.~\ref{fig:ablation}(d)), improving all metrics by effectively reducing bias in 2D edge images.
%
% Note, merging overlapping edges will cause metric drop bacause 

% Please add the following required packages to your document preamble:
% \usepackage{booktabs}
% \usepackage{multirow}
% \usepackage[normalem]{ulem}
% \useunder{\uline}{\ul}{}

\begin{table}[t]
\centering
\small
\begin{tabular}{l|cccccr}
\hline
Method                             & A$\downarrow$ & C$\downarrow$ & R5$\uparrow$  & P5$\uparrow$  & F5$\uparrow$ & E-num \\ \hline
\multirow{1}{*}{EdgeGS} 
                               & 7.4       & 7.2     & 75.7 & 86.9    & 80.3   & 140.76  \\
                               % \hline
\multirow{1}{*}{Ours (EdgeGS)} 
                               & \textbf{6.8}             &  \textbf{5.8}        & \textbf{90.8}    & \textbf{92.9} & \textbf{91.3}  & 44.28  \\ \hline
\multirow{1}{*}{EMAP}    
                               & 8.8    & 7.9     & 63.5 & 70.4    & 66.3  & 48.71        \\ % emap(2DGS-SN)
\multirow{1}{*}{Ours (EMAP)}    
                               & 6.9    & 6.2              & 88.8          & 90.8          & 89.3    & \textbf{31.70}        \\  % emap(2DGS-SN)
                               \hline
\end{tabular}
\caption{Ablation on initialization methods using ABC-NEF. \texttt{Ours (EdgeGS)} and  \texttt{Ours (EMAP)} denote our method initialized with EdgeGS and EMAP, respectively. 2DGS-SN is chosen as 2D edge detector.}
\label{tab:ablation-initialization}
\end{table}

\paragraph{Initialization Methods.}

In our main experiments, we use EdgeGS for initialization. However, we demonstrate SketchSplat can also achieve strong results when initialized with alternative methods like EMAP. As shown in Tab.~\ref{tab:ablation-initialization}, our approach consistently outperforms the initialization methods by a large margin and achieves comparable results across different initializations.

% \paragraph{Noisy Initialization}

% Keep robust for noisy initialization.
% \section{Limitations}

% \section{Discussion}

% Difference with EdgeGS: They optimize a set of needle-like gaussians, which are small lines locally and looks similar to our method. However, the there is no topology operations that can connect or merge them, and their final sketch fitting operation is not evaluated by multi-view images, which can be sensitive to local noise.

\section{Conclusion}
\label{sec:conclusion}

% We present SketchSplat, a method for parametric 3D edge reconstruction from calibrated multi-view images. Unlike previous approaches that rely on 3D edge point reconstruction and edge fitting, SketchSplat directly optimizes the parameters of curves and lines through differentiable sampling and multi-view sketch splatting, ensuring accurate alignment with input 2D images. A series of topological operations are proposed to adaptively refine the edge structure during training, which reduces redundancy while maintaining high accuracy and completeness. Additionally, an improved 2D edge detection method is adopted to achieve higher reconstruction quality. Experiments show that SketchSplat achieves the state-of-the-art accuracy and completeness on the ABC-NEF dataset while maintaining compact results and efficient training speed.

We present SketchSplat, a method for parametric 3D edge reconstruction from calibrated multi-view images. Unlike previous approaches that rely on 3D edge point reconstruction and edge fitting, SketchSplat directly optimizes edges through multi-view sketch splatting, ensuring accurate alignment with input 2D images. A series of topological operations are proposed to adaptively refine the edge structure during training, which reduces redundancy while maintaining high accuracy and completeness. Additionally, an improved 2D edge detection method is adopted to achieve higher reconstruction quality. Experiments show that SketchSplat achieves the state-of-the-art accuracy and completeness on the CAD dataset while maintaining compact results and efficient training speed.

\paragraph{Acknowledgements}
This research is based in part upon work supported by the Office of the Director of National  Intelligence (ODNI), Intelligence Advanced Research Projects Activity (IARPA), via IARPA R\&D Contract No. 140D0423C0076. The views and conclusions contained herein are those of the authors and should not be interpreted as necessarily representing the official policies or endorsements, either expressed or implied, of the ODNI, IARPA, or the U.S. Government. The U.S. Government is authorized to reproduce and distribute reprints for Governmental purposes notwithstanding any copyright annotation thereon.

{
    \small
    \bibliographystyle{ieeenat_fullname}
    \bibliography{main}
}

% \newpage
% \clearpage

%%%%%%%%% TITLE - PLEASE UPDATE
% \section*{Supplementary Material of \\SketchSplat: 3D Edge Reconstruction via Differentiable Multi-view Sketch Splatting}

\clearpage
\setcounter{section}{6}
\setcounter{table}{5}
\setcounter{figure}{4}

\maketitlesupplementary

\paragraph{Summary.}
In this supplementary material, we elaborate on the following topics:

\begin{itemize}
    \item Sec.~\ref{Details of 2D Edge Detector} provides details of our proposed edge detector 2DGS-SN, along with visual comparisons against existing 2D edge detectors.

    \item Sec.~\ref{Alternative 2D Edge Detector} presents an alternative depth estimator for 2D edge detection.

    \item Sec.~\ref{Robustness to Initialization} evaluates the robustness of our method under noisy initialization.

    \item Sec.~\ref{Necessity of Sketch Filtering} discusses the necessity of the sketch filtering operation.

    \item Metric definitions used in the main paper are detailed in Sec.~\ref{Details of Metrics}.

    \item Sec.~\ref{More Visual Results} presents additional qualitative results, including 20 scenes from the ABC-NEF dataset.

    \item In Sec.~\ref{Quantitative Evaluation on DTU dataset}, we report quantitative results on the DTU dataset.
    
    \item Sec.~\ref{Comparison to Feed-forward Baseline} compares our method with a feed-forward method.

    \item Finally, Sec.~\ref{Limitations} discusses the limitations of our approach.
\end{itemize}

\section{Details of 2D Edge Detector}\label{Details of 2D Edge Detector}
\begin{figure}[thb]
  \centering
  % \fbox{\rule{0pt}{2in} \rule{0.9\linewidth}{0pt}}
   \includegraphics[width=1.0\linewidth]{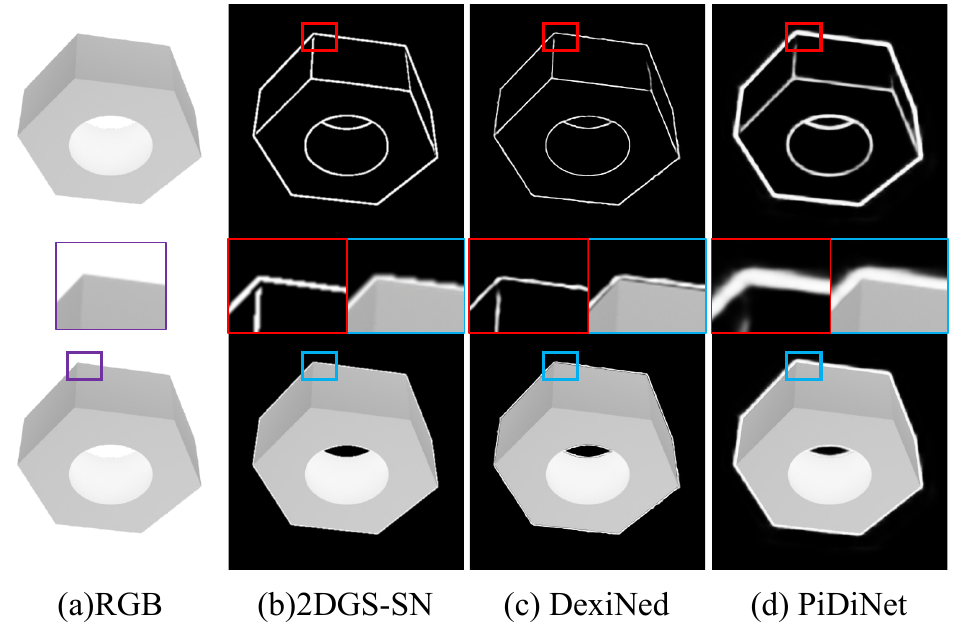}
   \caption{
   Visual comparison of different edge detection methods. Our 2DGS-SN provides more accurate edge detections results that align well with the object boundaries.
   }
   \label{fig:detect}
\end{figure}

We show the accuracy of 2DGS-SN by comparing with PiDiNet~\cite{su2021pidinet} and DexiNed~\cite{poma2020dexined}. From Fig.~\ref{fig:detect} (b), we can see 2DGS-SN produces accurate edges, where the object boundary locates at the center of the predicted edge pixels.
Both PiDiNet and DexiNed (Fig.~\ref{fig:detect} (c, d)) detect edges with slight offsets, positioning them far from the object boundaries. According to Tab.~1 in the main paper, this slight offset introduces significant multi-view inconsistency and ambiguity, resulting in degraded performance. This also explains why EdgeGS~\cite{chelani2024edgegs} achieve much higher F10 and F20 scores than EMAP~\cite{li2024emap}, but lower F5 scores.

\section{Alternative 2D Edge Detector}\label{Alternative 2D Edge Detector}
% 2DGS~\cite{huang20242dgs} provides depth maps with a multi-view consistent scale, so a single threshold $t_d$ in Eq.~3 is sufficient to detect consistent edges. 
The core insight of 2DGS-SN is that geometric cues (e.g., depth and normal) offer higher pixel-level accuracy and are thus more robust than pure learning-based edge detectors~\cite{poma2020dexined, su2021pidinet}.
Based on this observation, we demonstrate that other geometry estimation methods such as DepthPro~\cite{bochkovskii2024depthpro} can be alternative components of our 2DGS-SN. We replace 2DGS depth maps with the ones from DepthPro and test its performance. DepthPro is significantly faster than 2DGS, as it avoids per-scene optimization. However, its accuracy is lower due to unstable edge detection caused by multi-view inconsistent depth scales (see Tab.~\ref{tab:ablation-2d-detector-depth}). In contrast,
2DGS~\cite{huang20242dgs} provides depth maps with a multi-view consistent scale, therefore a single threshold $t_d$ in Eq.~3 is sufficient
to detect consistent edges.
Choosing 2DGS (higher accuracy) vs. DepthPro (faster) is therefore a trade-off between efficiency and quality.

% Please add the following required packages to your document preamble:
% \usepackage{booktabs}
% \usepackage{multirow}
% \usepackage[normalem]{ulem}
% \useunder{\uline}{\ul}{}

\begin{table}[t]
\centering
\footnotesize
\begin{tabular}{l|c|ccccc}
\hline
Method                           & Detector  & A$\downarrow$ & C$\downarrow$ & R5$\uparrow$  & P5$\uparrow$  & F5$\uparrow$ \\ \hline
\multirow{1}{*}{EdgeGS} & 2DGS-SN
                               & 7.4       & 7.2     & 75.7 & 86.9    & 80.3  \\
                               % \hline
\multirow{1}{*}{Ours} & 2DGS-SN
                               & \textbf{6.8}             &  5.8        & \textbf{90.8}    & \textbf{92.9} & \textbf{91.3}  \\ \hline
\multirow{1}{*}{EdgeGS} & DepthPro-SN
                               & 10.6    & 7.0              & 75.5          & 81.8          & 77.9        \\ 
\multirow{1}{*}{Ours} & DepthPro-SN
                               & 9.2    & \textbf{5.4}              & 90.3          & 87.5          & 87.8       \\ \hline
                               
\end{tabular}
\caption{\footnotesize \textbf{Ablation on 2D edge detector}. We ablate the depth prediction methods (2DGS~\cite{huang20242dgs}, DepthPro~\cite{bochkovskii2024depthpro}). 2DGS shows better performance since it produces multi-view consistent depth maps. All experiments use EdgeGS as initialization method.}
\label{tab:ablation-2d-detector-depth}
\end{table}

% \section{Hyperparameter Generalizability}
% The percentage and angular thresholds (e.g., $th_{overlap}$ and $th_{dir}$) are general parameters that work across different scenes without adjustment.
% Distance thresholds (e.g., $th_{connect}$) are set based on the desired precision and we set all of them as $1\%$ of the scene size ($1 m^3$).

\section{Robustness to Initialization}\label{Robustness to Initialization}
\begin{figure*}[thb]
  \centering
  % \fbox{\rule{0pt}{2in} \rule{0.9\linewidth}{0pt}}
   \includegraphics[width=0.8\linewidth]{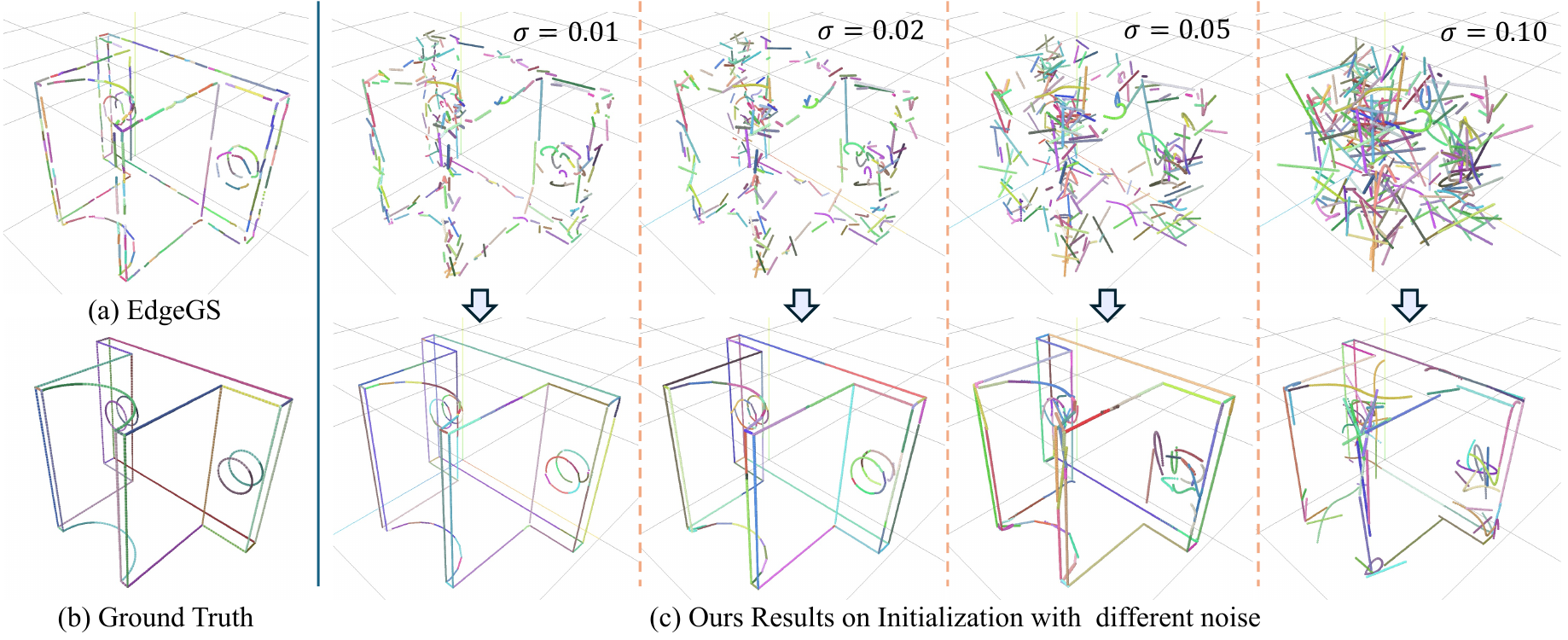}
   \caption{
   \textbf{Performance under noisy initialization}. We add Gaussian noise $\delta\sim\mathcal{N}(0,\sigma^2)$ to our initialization edges (from EdgeGS). On ABC-NEF, our method remains robust within reasonable noise levels but degrades when initial edge lengths become excessively large relative to the scene size $1 m^3 (\sigma>0.02)$, which is rare in our experiments. 
   }
   \label{fig:noisy_init}
\end{figure*}

To illustrate the robustness of our method, we inject Gaussian noise $\delta\sim\mathcal{N}(0,\sigma^2)$ to the parameters of the edges initialized from EdgeGS. Fig.~\ref{fig:noisy_init} and Tab.~\ref{tab:ablation-initialization-noise} show that 
% our method performs well even with noisy initializations but degrades when initial edge lengths become excessively large relative to the scene size, which is rare in our experiments. 
our method performs robust even with noisy initializations and only degrades when initial edge lengths become excessively large relative to the scene size, which rarely occurs in practice.
% Pure random initialization as suggested by \Ra~is an interesting topic for future work.
Exploring alternative initialization methods would be an interesting topic for future work.

% Please add the following required packages to your document preamble:
% \usepackage{booktabs}
% \usepackage{multirow}
% \usepackage[normalem]{ulem}
% \useunder{\uline}{\ul}{}

% \footnotesize
% \scriptsize

\begin{table}[t]
\centering
\footnotesize
\begin{tabular}{l|c|ccccc}
\hline
Method                           & Noise ($\sigma$)  & A$\downarrow$ & C$\downarrow$ & R5$\uparrow$  & P5$\uparrow$  & F5$\uparrow$ \\ \hline

\multirow{1}{*}{EdgeGS} & -
                               & 7.4       & 7.2     & 75.7 & 86.9    & 80.3  \\
                               % \hline
\multirow{1}{*}{Ours} & -
                               & 6.8             &  5.8        & 90.8    & 92.9 & 91.3  \\ \hline
\multirow{1}{*}{Ours} & 0.01
                               & 6.7    & 5.7              & 91.0          & 92.9          & 91.4        \\ 
\multirow{1}{*}{Ours} & 0.02
                               & 6.6    & 6.2              & 91.0          & 92.4          & 91.2       \\ 
\multirow{1}{*}{Ours} & 0.05
                               & 7.4    & 8.4              & 87.3          & 84.3          & 85.2       \\ 
\multirow{1}{*}{Ours} & 0.10
                               & 21.4    & 17.5              & 65.8          & 54.1          & 58.3       \\ 
                               \hline
                               
\end{tabular}
\caption{\footnotesize \textbf{Performance under Noisy Initialization}. We present results with noisy initialization by adding Gaussian noise to the input edge parameters (on ABC-NEF dataset). Compared to the baseline EdgeGS, our method performs significantly better within a reasonable noise level ($\sigma < 0.05$ for a $1m^3$ scene). All results use 2DGS-SN as edge detector.}
\label{tab:ablation-initialization-noise}
\end{table}

\section{Necessity of Sketch Filtering}\label{Necessity of Sketch Filtering}
In the main paper, the proposed filtering operations seem to have little effect on the metrics (Tab.~3), as the edges initialized from EdgeGS are already clean. However, under poor initialization, filtering becomes essential for improving accuracy. We demonstrate this using a pseudo noisy initialization scenario, where we add Gaussian noise to the initialized edge parameters with $\sigma=0.02$ and examine the effect of filtering. Tab.~\ref{tab:ablation-filter-noise} shows that without filtering, the method suffers a noticeable accuracy drop and slightly higher completeness, confirming our assumption.

% # no filtering - 0.02 noise
% # recall_0.005: 0.9103316972462948
% # precision_0.005: 0.907564654225226
% # fscore_0.005: 0.9036680516471768
% # IOU_0.005: 0.33451337120864966
% # chamfer_dist: 0.012903088703751564
% # acc: 0.007048003375530243
% # comp: 0.0058550843968987465

% Please add the following required packages to your document preamble:
% \usepackage{booktabs}
% \usepackage{multirow}
% \usepackage[normalem]{ulem}
% \useunder{\uline}{\ul}{}

% \footnotesize
% \scriptsize

\begin{table}[t]
\centering
\footnotesize
\begin{tabular}{l|c|c|ccccc}
\hline
Method                           & Noise ($\sigma$) & Topo  & A$\downarrow$ & C$\downarrow$ & R5$\uparrow$  & P5$\uparrow$  & F5$\uparrow$ \\ \hline

\multirow{1}{*}{Ours} & - & all
                               & 6.8             &  5.8        & 90.8    & 92.9 & 91.3  \\ \hline

\multirow{1}{*}{Ours} & 0.02 & all
                               & 6.6    & 6.2              & 91.0          & 92.4          & 91.2       \\ 
\multirow{1}{*}{Ours} & 0.02 & no filter
                               & 7.0    & 5.9              & 91.0          & 90.7          & 90.4       \\ 
                               
                               \hline
                               
\end{tabular}
\caption{\footnotesize \textbf{Effectiveness of Filtering Operations under Noisy Initialization}. We evaluate performance under noisy initialization by adding Gaussian noise to the input edge parameters on the ABC-NEF dataset. Without filtering, the method shows a clear drop in accuracy.}
\label{tab:ablation-filter-noise}
\end{table}

% Please add the following required packages to your document preamble:
% \usepackage{booktabs}
% \usepackage{multirow}
% \usepackage[normalem]{ulem}
% \useunder{\uline}{\ul}{}
\begin{table*}[thb]
\centering
\small
\begin{tabular}{c|cc|cc|cc|cc|cc|cc|cc}
\hline
\multirow{2}{*}{Scan} & \multicolumn{2}{c|}{LIMAP~\cite{liu2023limap}}   & \multicolumn{2}{c|}{NEF~\cite{ye2023nef}}    & \multicolumn{2}{c|}{NEAT~\cite{xue2024neat}}    & \multicolumn{2}{c|}{EMAP~\cite{li2024emap}}     & \multicolumn{2}{c|}{EdgeGS*~\cite{chelani2024edgegs}} & \multicolumn{2}{c|}{Ours} & \multicolumn{2}{c}{Ours (Topo)} \\
                      & R5$\uparrow$  & P5$\uparrow$ & R5$\uparrow$ & P5$\uparrow$ & R5$\uparrow$ & P5$\uparrow$  & R5$\uparrow$  & P5$\uparrow$  & R5$\uparrow$  & P5$\uparrow$ & R5$\uparrow$      & P5$\uparrow$     & R5$\uparrow$     & P5$\uparrow$    \\ \hline
37                    & 75.8          & 74.3         & 39.5         & 51.0         & 63.9         & \textbf{85.1} & 62.7          & 83.9          & \textbf{79.4} & 76.6         & 78.7              & 80.2             & 78.2             & 78.9            \\
83                    & 75.7          & 50.7         & 32.0         & 21.8         & 72.3         & 52.4          & 72.3          & 61.5          & \textbf{77.8} & 62.6         & 76.5              & \textbf{64.0}    & 74.2             & 62.8            \\
105                   & \textbf{79.1} & 64.9         & 30.3         & 32.0         & 68.9         & 73.3          & 78.5          & \textbf{78.0} & 72.6          & 68.8         & 74.0              & 72.0             & 71.6             & 72.3            \\
110                   & 79.7          & 65.3         & 31.2         & 40.2         & 64.3         & \textbf{79.6} & \textbf{90.9} & 68.3          & 83.4          & 60.3         & 85.0              & 66.8             & 83.5             & 66.1            \\
118                   & 59.4          & 62.0         & 15.3         & 25.2         & 59.0         & 71.1          & \textbf{75.3} & \textbf{78.1} & 74.5          & 68.6         & 73.5              & 70.5             & 71.2             & 70.1            \\
122                   & 79.9          & 79.2         & 15.1         & 29.1         & 70.0         & 82.0          & 85.3          & 82.9          & 85.0          & 82.7         & \textbf{86.8}     & \textbf{83.0}    & 83.6             & 81.9            \\ \hline
mean                  & 74.9          & 66.1         & 27.2         & 33.2         & 66.4         & 73.9          & 77.5          & \textbf{75.4} & 78.8          & 70.0         & \textbf{79.1}     & 72.8             & 77.0             & 72.0            \\ \hline
\end{tabular}
\caption{3D Edge Reconstruction on the DTU dataset~\cite{aanaes2016dtu}. Our SketchSplat performs on par with EMAP~\cite{li2024emap} and EdgeGS*~\cite{chelani2024edgegs} (EdgeGS* denotes the results reproduced with the code released by the authors~\cite{chelani2024edgegs}). \haiyang{Ours(Topo): ours with topology control.}
}
\label{tab:dtu}
\end{table*}

\section{Details of Metrics} \label{Details of Metrics}

We provide the details of the used metrics in experiments. 
The ABC-NEF dataset~\cite{koch2019abc, ye2023nef} provides ground-truth edges, which can be used for quantitative evaluation of 3D parametric edge reconstruction. 
To compute the metrics, we sample points along the predicted edges and compare them with points sampled at the same resolution from the ground-truth edges.

Accuracy (A) is defined as the average distance from predicted points to their nearest ground-truth points, while completeness (C) measures the average distance from ground-truth points to the closest predicted points. Lower values indicate better performance for these metrics.

For a given distance threshold $\tau$, precision P($\tau$) represents the proportion of predicted points with at least one corresponding ground-truth point within $\tau$. 
Conversely, recall R($\tau$) indicates the percentage of ground-truth points that have a predicted counterpart within the same threshold. The F-score F($\tau$) is the harmonic mean of precision and recall, formulated as:
\begin{equation}
    F(\tau) = \frac{2P(\tau)R(\tau)}{P(\tau) + R(\tau)}
\end{equation}
For precision, recall, and F-score, higher values indicate better performance. We report these metrics for thresholds of 5, 10, and 20 millimeters (mm). All of the CAD objects are scaled such that the longest edge of the bounding box is normalized to 1 meter (m).

\section{More Visual Results}\label{More Visual Results}

\begin{figure*}[thb]
  \centering
   \includegraphics[width=1.0\linewidth]{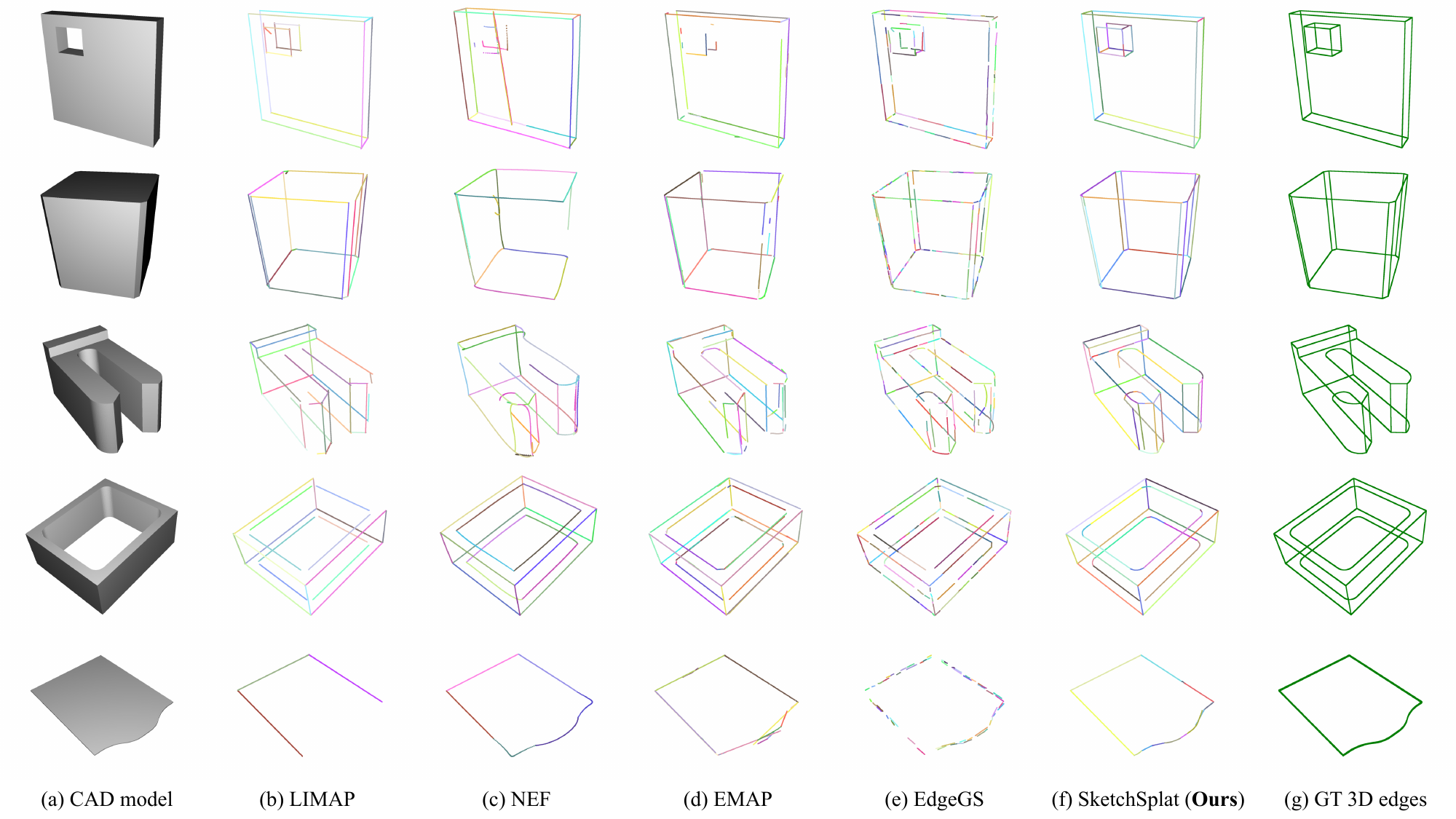}
   \caption{More qualitative results on ABC-NEF dataset~\cite{ye2023nef}.}
   \label{fig:qua2}
\end{figure*}

\begin{figure*}[thb]
  \centering
   \includegraphics[width=1.0\linewidth]{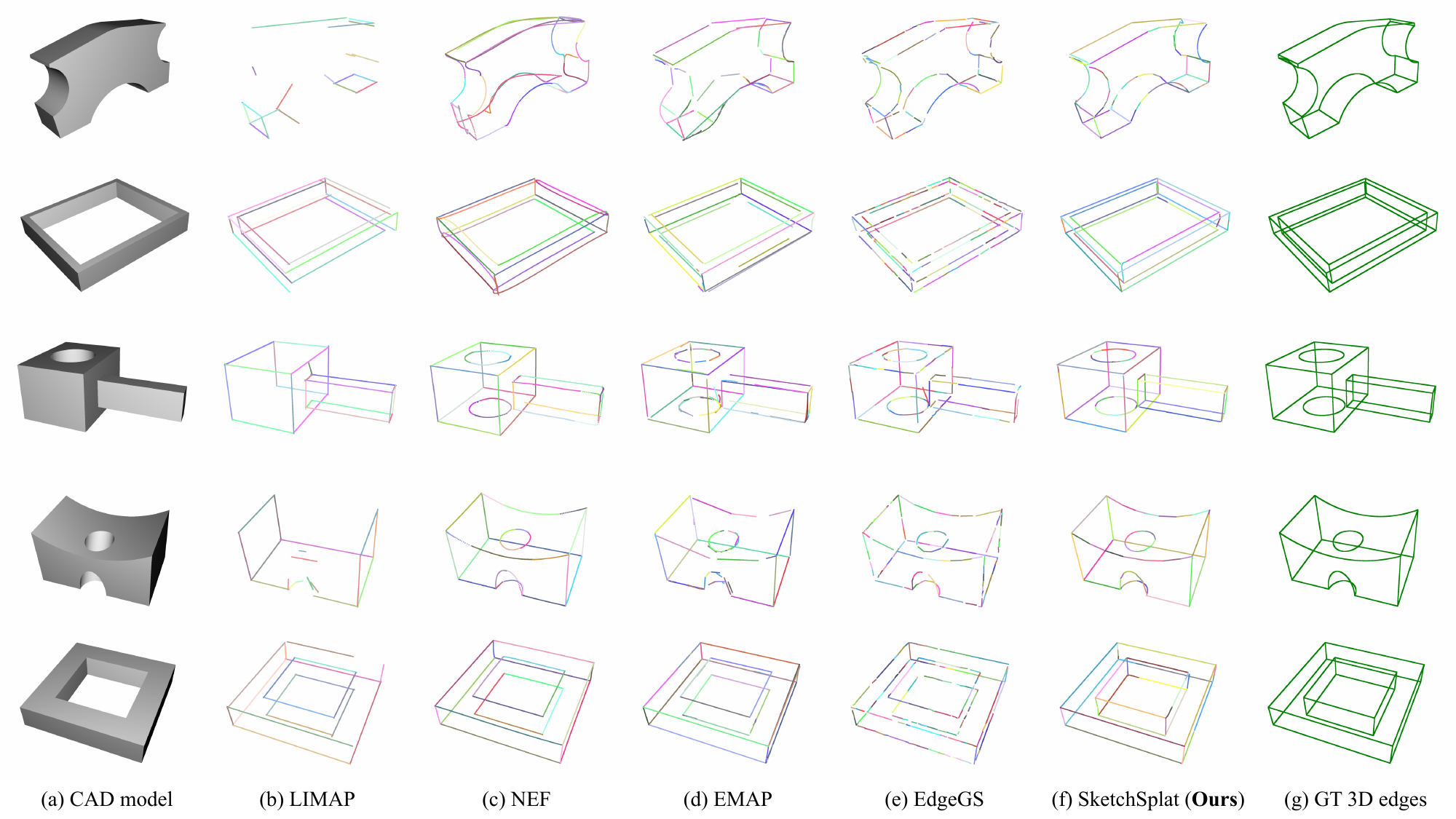}
   \caption{More qualitative results on ABC-NEF dataset~\cite{ye2023nef}.}
   \label{fig:qua3}
\end{figure*}

\begin{figure*}[thb]
  \centering
   \includegraphics[width=1.0\linewidth]{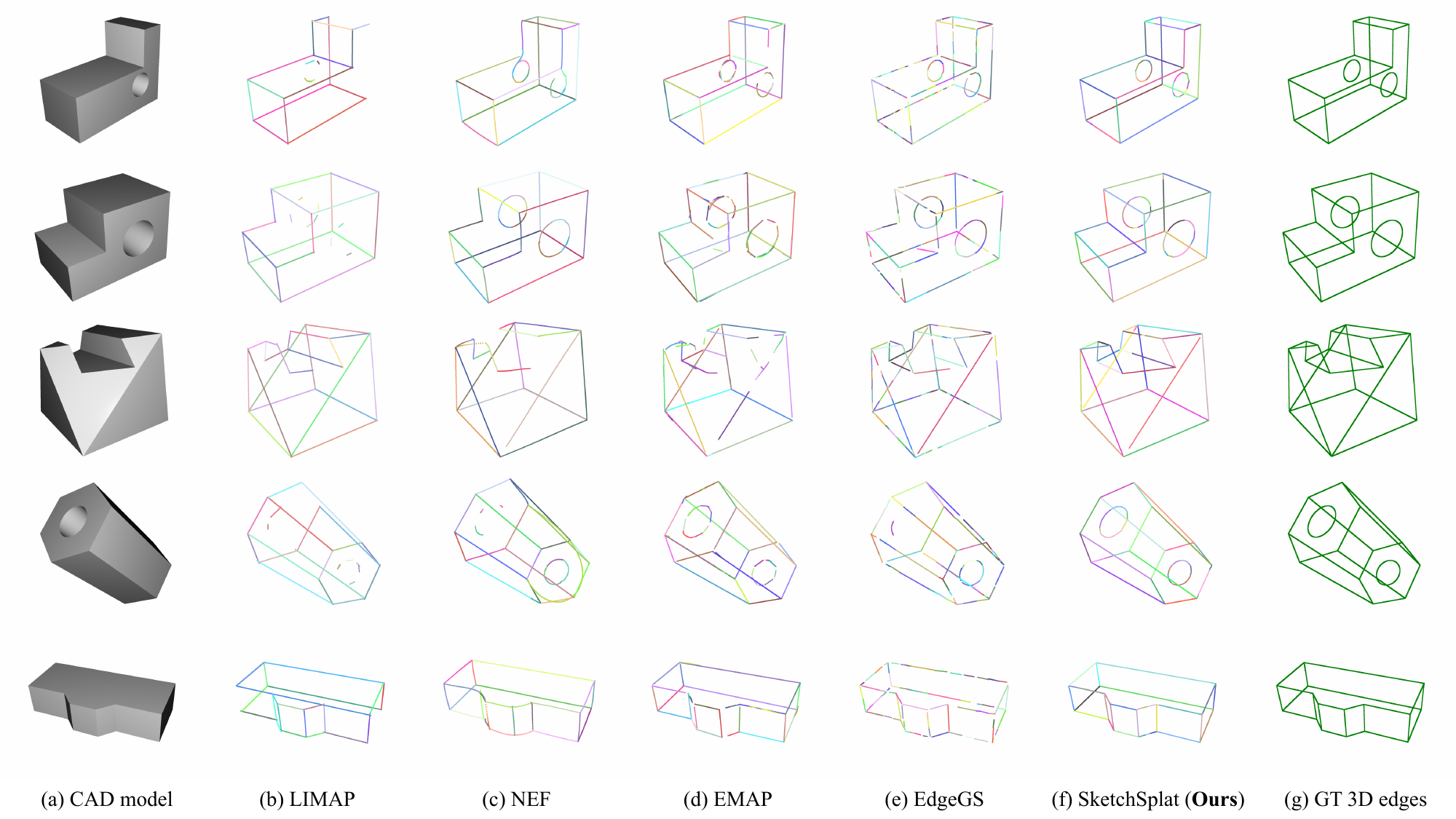}
   \caption{More qualitative results on ABC-NEF dataset~\cite{ye2023nef}.}
   \label{fig:qua4}
\end{figure*}

\begin{figure*}[thb]
  \centering
   \includegraphics[width=1.0\linewidth]{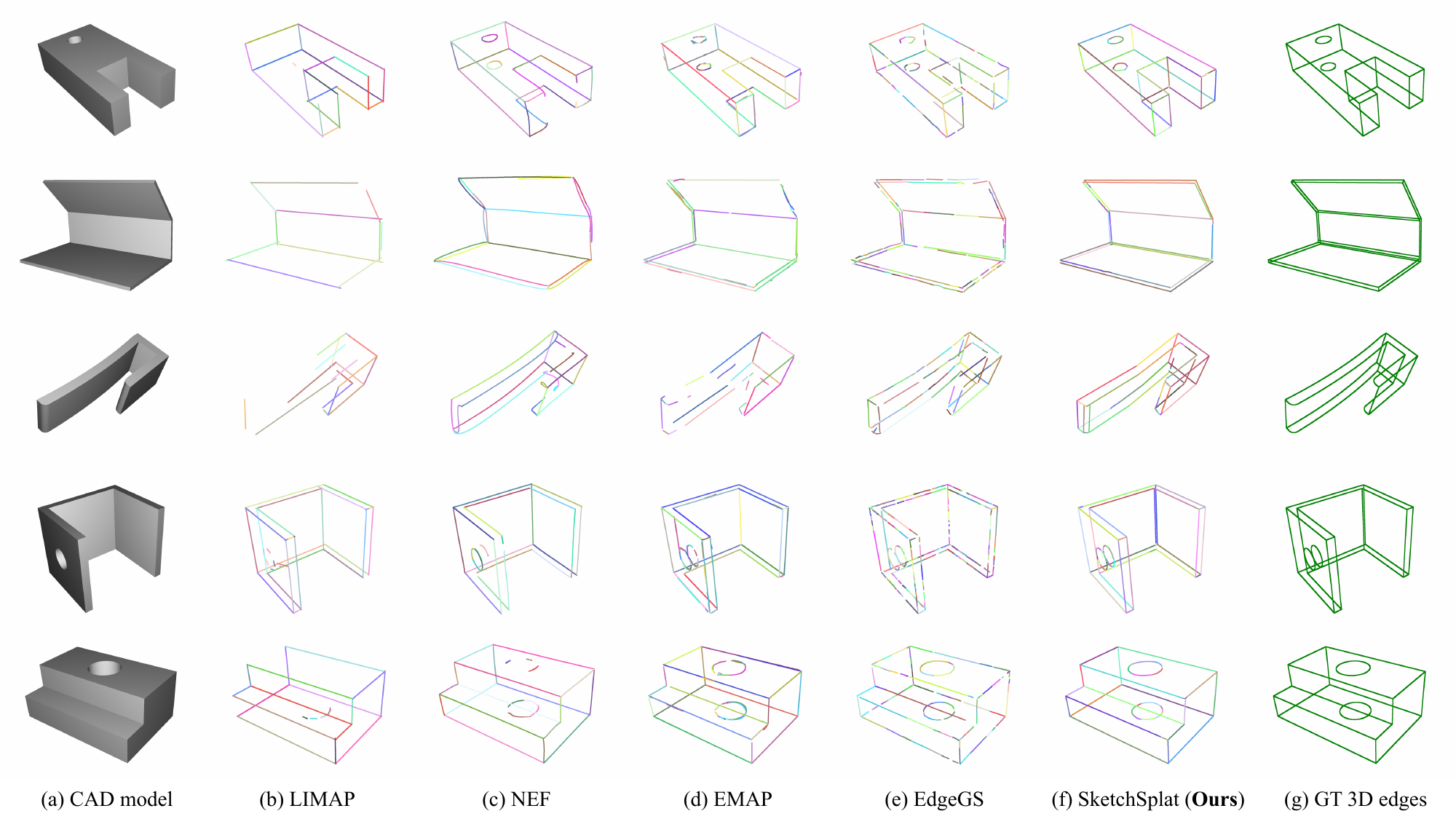}
   \caption{More qualitative results on ABC-NEF dataset~\cite{ye2023nef}.}
   \label{fig:qua5}
\end{figure*}

We provide more visualization results (20 scenes) on ABC-NEF dataset~\cite{ye2023nef, koch2019abc} in Fig.~\ref{fig:qua2}, Fig.~\ref{fig:qua3}, Fig.~\ref{fig:qua4}, and Fig.~\ref{fig:qua5}. These figures show that our method achieves state-of-the-art accuracy and completeness compared to existing approaches.

\section{Quantitative Evaluation on DTU dataset} \label{Quantitative Evaluation on DTU dataset}
Tab.~\ref{tab:dtu} shows the evaluation results on six scenes in the DTU dataset~\cite{aanaes2016dtu}. We reproduce EdgeGS~\cite{chelani2024edgegs} using its released code and configuration (denoted as EdgeGS*) to initialize sketches for our method. 
As shown in Tab.~\ref{tab:dtu}, SketchSplat without topological operations achieves comparable results to EMAP~\cite{li2024emap} and slightly outperforms EdgeGS, demonstrating its effectiveness in correcting misalignment between 3D edges and 2D edge images.
However, incorporating topological operations slightly reduces recall, as merging reduces redundant edges, whereas the pseudo ground-truth favors duplicated edges in P5 and R5 due to the thickness of the GT annotations (see Fig.~4 in the main paper).

\section{Comparison to Feed-forward Baseline} \label{Comparison to Feed-forward Baseline}
We also compare against a feed-forward 3D line detection method. To evaluate methods like NerVE~\cite{zhu2023nerve}, we first reconstruct a point cloud from the input images (e.g., using 2DGS), then input it to the pretrained NerVE model to predict 3D edges. As shown in Tab.~\ref{tab:ablation-feedforward-method}, NerVE performs significantly worse than our method and the baselines, suggesting that optimization-based methods remain more robust and generalize better, while feed-forward approaches are more sensitive to noisy inputs.

% Please add the following required packages to your document preamble:
% \usepackage{booktabs}
% \usepackage{multirow}
% \usepackage[normalem]{ulem}
% \useunder{\uline}{\ul}{}

\begin{table}[t]
\centering
\footnotesize
\begin{tabular}{l|c|ccccc}
\hline
Method                           & Input  & A$\downarrow$ & C$\downarrow$ & R5$\uparrow$  & P5$\uparrow$  & F5$\uparrow$ \\ \hline
\multirow{1}{*}{EMAP}   & 2DGS-SN
                               & 8.8       & 7.9     & 63.5 & 70.4    & 66.3  \\ \hline
\multirow{1}{*}{EdgeGS} & 2DGS-SN
                               & 7.4       & 7.2     & 75.7 & 86.9    & 80.3  \\ \hline
\multirow{1}{*}{Ours}   & 2DGS-SN
                               & \textbf{6.8}    & \textbf{5.8}              & \textbf{90.8}          & \textbf{92.9}          & \textbf{91.3}       \\ \hline
\multirow{1}{*}{NerVE (PWL)}   & 2DGS
                               & 18.0    & 35.7              & 40.5          & 62.3          & 48.4        \\ 
\multirow{1}{*}{NerVE (CAD)}   & 2DGS
                               & 15.3    & 117.3              & 24.0          & 57.7          & 32.3        \\ 
                               \hline
                               
\end{tabular}
\caption{\textbf{Comparison with feed-forward method NerVE on ABC-NEF dataset}. For NerVE, we train a 2DGS and extract surface points as input. It turns out that NerVE is sensitive to noise in the input and produces incomplete and inaccurate edges.
% NerVE (CAD)*: only 43 out of 82 scenes are valid. 
Our method use EdgeGS as initialization.}
\label{tab:ablation-feedforward-method}
\end{table}

\section{Limitations}\label{Limitations}

\begin{figure*}[thb]
  \centering
   \includegraphics[width=1.0\linewidth]{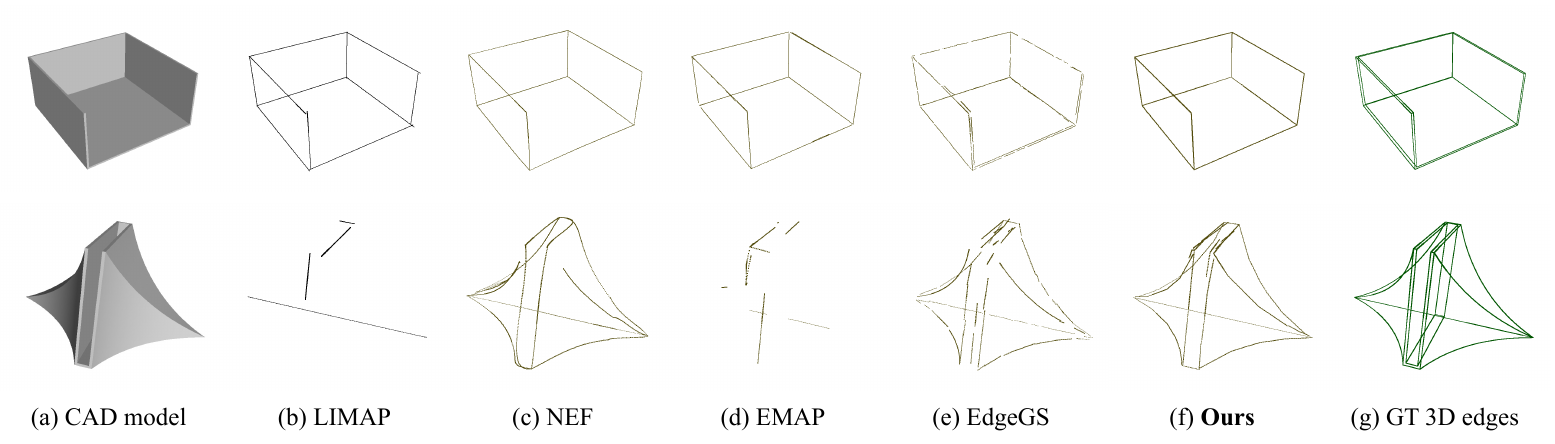}
   \caption{Failure cases. Our method struggles with very thin structures and the wrong edges.}
   \label{fig:limitation}
\end{figure*}

\paragraph{Objects with thin structures.} Though our method performs well on small structure (like scene 7551 in ABC-NEF dataset), our merge mechanism may fail when edges with distance close to or lower than our merging threshold (such as the case shown in the first raw of Fig.~\ref{fig:limitation}).

\paragraph{Wrong supervision.} Like previous methods~\cite{ye2023nef, li2024emap, chelani2024edgegs}, our approach also encounters the challenge of incorrect edge supervision. During rendering, both visible and invisible edges from a given view are projected onto 2D images. As a result, some invisible edges may receive positive supervision, leading to misplaced edges that are difficult to filter out. This issue is illustrated in the second row of Fig.~\ref{fig:limitation}.

\paragraph{Edge detection} Although the proposed edge detector 2DGS-SN is accurate, we found when both depth reconstruction and normal estimation fail, some small structure such as small circles will not be detected in 2D edge images. This leads to edge missing in the reconstructed 3D edges. Further improving the edge detection method may resolve this problem.

\end{document}